%% file: main.tex
\documentclass[pdftex,11pt,openright,headsepline]{book}
\usepackage[margin=1in,headheight=13.6pt]{geometry}

\usepackage{paralist}		
\usepackage{color}		
\usepackage{times}
\usepackage{amsfonts}		
\usepackage{amsmath}		
\usepackage{latexsym}
\usepackage{graphicx}		
\usepackage{mydefs}		
\usepackage{fancyhdr}		
\input{shortcuts.tex}

\begin{document}
\pagestyle{empty} 

\fancypagestyle{plain}{
  \renewcommand{\headrulewidth}{0.0pt}
  \fancyfoot{}
  \fancyhead{}
}

\input{title.tex}

\cleardoublepage

\input{abstract.tex}

\cleardoublepage
\newpage

\fancypagestyle{plain}{
  \renewcommand{\headrulewidth}{0.0pt}
  \fancyfoot{}
  \fancyfoot[RO, LE]{\thepage}
  \fancyhead{}
}

\pagestyle{fancy}
\pagenumbering{Roman}

\tableofcontents

\listoffigures
\cleardoublepage

\listoftables
\cleardoublepage

\newpage

\pagenumbering{arabic}

\input{text/1-intro}
\input{text/2-relatedwork}
\input{text/3-materialsandmethods}
\input{text/4-experimentsandresults}
\input{text/5-conclusion}

\appendix
\input{appendix/appendix-A}
\input{appendix/appendix-B}
\input{appendix/appendix-C}
\input{appendix/appendix-D}

%
%

\bibliographystyle{plain}
\bibliography{references}

\end{document}

%% file: shortcuts.tex
\newcommand\mypara[1]{\vspace{1mm}\noindent\textbf{#1}}
\usepackage{algpseudocode}
\usepackage{algorithm}

%% file: title.tex
%

\begin{titlepage}

\thispagestyle{empty}

\fancypagestyle{empty}{
\lhead{\includegraphics[height=1.5cm]{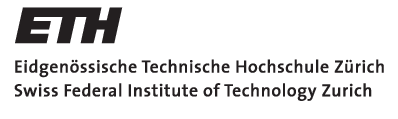}}
\renewcommand{\headrulewidth}{0.0pt}
\rhead{\vspace*{-0.2cm}\includegraphics[height=1.4cm]{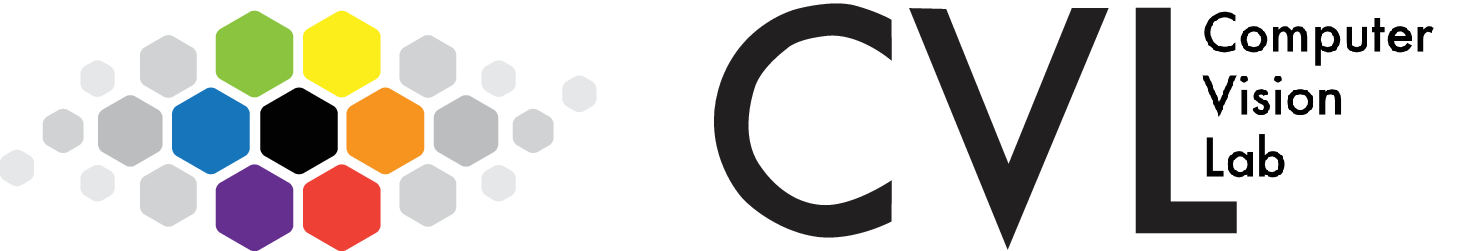}}
\fancyfoot{}
}

\vspace*{2cm}
\begin{center}
\LARGE{\textbf{NPSim: Nighttime Photorealistic Simulation From Daytime Images With Monocular Inverse Rendering and Ray Tracing
}\\}
\vspace{5pt}
\large{Project Thesis\\[0.8cm]}
\LARGE{Shutong Zhang\\}
\normalsize{Department of Information Technology and Electrical Engineering}
\end{center}

\begin{center}


\end{center}

\vfill
\begin{center}
\begin{tabular}{ll}
\Large{\textbf Advisor:} & \Large{Dr.~Christos Sakaridis}\\
\Large{\textbf Supervisor:} & \Large{Prof.~Dr.~Luc Van Gool}\\
\end{tabular}
\end{center}

\begin{center}
August 18, 2023
\end{center}

\end{titlepage}

%% file: abstract.tex
%

\newpage
\vspace{3cm}

\chapter*{Abstract}



Semantic segmentation is an important task for autonomous driving. A powerful autonomous driving system should be capable of handling images under all conditions, including nighttime. Generating accurate and diverse nighttime semantic segmentation datasets is crucial for enhancing the performance of computer vision algorithms in low-light conditions. In this thesis, we introduce a novel approach named NPSim, which enables the simulation of realistic nighttime images from real daytime counterparts with monocular inverse rendering and ray tracing. NPSim comprises two key components: mesh reconstruction and relighting. The mesh reconstruction component generates an accurate representation of the scene’s structure by combining geometric information extracted from the input RGB image and semantic information from its corresponding semantic labels. The relighting component integrates real-world nighttime light sources and material characteristics to simulate the complex interplay of light and object surfaces under low-light conditions. The scope of this thesis mainly focuses on the implementation and evaluation of the mesh reconstruction component. Through experiments, we demonstrate the effectiveness of the mesh reconstruction component in producing high-quality scene meshes and their generality across different autonomous driving datasets. We also propose a detailed experiment plan for evaluating the entire pipeline, including both quantitative metrics in training state-of-the-art supervised and unsupervised semantic segmentation approaches and human perceptual studies, aiming to indicate the capability of our approach to generate realistic nighttime images and the value of our dataset in steering future progress in the field. NPSim not only has the ability to address the scarcity of nighttime datasets for semantic segmentation, but it also has the potential to improve the robustness and performance of vision algorithms under low-lighting conditions.

%% file: text/1-intro.tex
%

\chapter{Introduction}




Accurately parsing input images under uncommon visual conditions is a required ability for safe autonomous driving systems. Thus, semantic segmentation datasets under various adverse conditions are required, especially at nighttime. However, most well-known autonomous driving datasets such as the Cityscapes dataset~\cite{Cordts2016Cityscapes} and KITTI dataset~\cite{Geiger2012CVPR} are mostly under normal conditions (clear daytime image with high illumination and low exposure time). Though several recent works, including Oxford RobotCar dataset~\cite{RCDRTKArXiv}, BDD100K~\cite{bdd100k} dataset and ACDC dataset~\cite{SDV21ACDC}, have been focusing on creating datasets under nighttime conditions, there are still obvious limitations and gaps. For example, the Oxford RobotCar does not contain any semantic annotations and thus is not capable for training a segmentation network. BDD100K dataset contains a large number of nighttime images; however, only 345 of them can be used for the task of semantic segmentation. ACDC dataset contains 4006 adverse condition images in which 1006 of them are in the nighttime; it still shows a drastic drop in the accuracy of semantic scene understanding at nighttime compared to other conditions for both segmentation algorithms~\cite{2016_deeplab, deeplabv3plus2018, fu2019danet, Lin_2017_RefineNet, 2015_SFSU, SunXLW19} and domain adaptation methods~\cite{2019_bdl, luo2019Taking, SDHV18, Tsai_adaptseg_2018, vu2018advent, Wang_2020_CVPR, 2020_fda}. What makes things even worse is the increased difficulty in the semantic annotation of real nighttime images due to their low quality, which leads to annotation errors that have a negative impact on the quality of models trained on such data. An alternative approach for semantic nighttime scene understanding is via the generation of partially synthetic nighttime data. Specifically, individuals can obtain images captured during the daytime, which is less noisy and relatively easier to annotate, then transform these images to nighttime through style transfer~\cite{2015_style, 2016_style}. Subsequently, annotations on daytime images can be directly used to the synthesized nighttime images, given that the underlying semantic content remains consistent. However, style transfer falls short in generating realistic results due to its inability to account for factors such as changes in illumination that occur from day to night and the complicated geometry and variations in light sources. In this thesis, we propose the NPSim, an alternative approach for generating partially synthetic nighttime images via monocular inverse rendering and ray tracing. 

Our proposed method NPSim aims to generate photo-realistic nighttime images based on clear daytime images and their corresponding standard semantic annotations. Importantly, we also utilize light source annotations as additional input. Different from previous works that used style transformation~\cite{CycleGAN2017}, 2D~\cite{Punnappurath_2022_CVPR} or implicit representation such as NeRF~\cite{mildenhall2020nerf}, NPSim focuses on the fundamentals of scene lighting, using traditional rendering method ray tracing to produce realistic nighttime images by restoring the object orientation, material characteristics, and light sources based on the input image. It leverages the explicit representation by reconstructing scene mesh from a given RGB image. To preserve more accurate geometric information, we propose Geometry Mesh Reconstruction component: We first utilize an off-the-shelf depth and normal estimation model to predict the initial depth map and normal map. Then we use predicted depth to reconstruct the scene mesh and refine it using a normal-guided optimization-based method. Besides, our proposed NPSim considers real-world light sources to generate realistic nighttime images. In the Realistic Nighttime Scene Relighting component, we employ ray tracing to generate nighttime visuals by considering geometric scene mesh from previous steps, material attributes, and authentic light sources from the real world. Our probabilistic light source activation also has the potential to activate different light source combinations, generating multiple nighttime images based on one single input image.

In summary, the contributions of our work are:
\begin{itemize}
\setlength{\itemsep}{1mm}
\setlength{\parskip}{0pt}
    \item A dataset of inactive light sources and the corresponding light source dataset, containing strength and chromaticity tuples (\mysecref{sec:data_prep}).
    \item A physically-based day-to-night simulation pipeline that contains a Geometry Mesh Reconstruction component (\mysecref{sec:gmr}, \mysecref{sec:drk}, \mysecref{sec:mpk}) and a Realistic Nighttime Scene Relighting component (\mysecref{sec:rnsr}).
\end{itemize}
In the experiment, our method can achieve better geometry reconstruction compared with previous works~\cite{hu2021worldsheet, zhang2022simbar}. Additionally, we show that our method can be generalized to other autonomous driving datasets such as the Cityscapes dataset~\cite{Cordts2016Cityscapes} and the BDD100K dataset~\cite{bdd100k} for the task of mesh reconstruction. The structure of the thesis is as follows: We first describe methods we used for data collection, then we introduce our day-to-night transformation method NPSim. In the end, we present the results of the Geometric Mesh Reconstruction component and testing plans for the entire data generation pipeline.

%% file: text/2-relatedwork.tex
%
\newpage
\chapter{Related Work}



Our work aims to create realistic nighttime images for autonomous driving based on single-view daytime images, it is closely related to novel view synthesis, day-to-night transformation and nighttime driving scene understanding. In this section, we will present an overview of the most relevant works and their limitations.

\section{Novel View Synthesis}
View synthesis is a fundamental task in computer vision that involves generating new images of a scene from different viewpoints. Early works focused on geometric reconstruction usually combine Structure-from-Motion (SfM) and Multi-View Stereo (MVS) that rely on sparse feature matching and depth estimation. However, these methods require multiple viewpoints of a single scene and cannot handle complex scenes~\cite{2006_MVS}. The recent influential Neural Radiance Fields (NeRF) technique~\cite{mildenhall2020nerf} shows strong ability in novel view synthesis and is capable for both indoor and outdoor scenes~\cite{mine2021iccv, lyu2022nrtf, srinivasan2021cvpr, rudnev2022nerfosr} or objects~\cite{nerv2021}. Different from previous works that leverage geometry information, it often designs as a multi-layer perceptron (MLP)~\cite{popescu2009mlp} that maps 3D coordinates to radiance and density values. NeRF learns to model the radiance field by minimizing the difference between the synthesized images and the actual training image during training, then renders different views of the scene at test time. Though being powerful and reliable in view synthesis, NeRF suffers from its high computational cost and limited generalizability. Many other works also explored view synthesis using generative networks, such as Variational Autoencoders (VAEs)~\cite{kingma2014vae}, Generative Adversarial Networks (GANs)~\cite{goodfellow2020gan} and Diffusion Models~\cite{ho2020denoising}. They have shown remarkable abilities in generating realistic images, but their lack of 3D understanding makes it hard to capture the underlying geometry and thus may generate artifacts in novel views.

\section{Day-to-Night Transformation}
Day-to-night transformation is a challenging task that aims to convert images captured during the day into realistic nighttime representations. This process relies on scene relighting, a core task in computer graphics and computer vision that involves modifying the lighting conditions and then rendering the original scene under new conditions. Many previous works have explored this with different methodologies. \cite{li2020inverse} learns inverse rendering from a single image, estimating the geometry and materials of the scene and spatially-varying illumination. \cite{Yang_2023_CVPR} proposed to complement the intrinsic estimation from volume rendering using NeRF and from inversing the photometric image formation model using convolutional neural networks (CNNs) for outdoor scene relighting. Differently, \cite{zhang2022simbar} leveraging explicit geometric representations from a single image by estimating depth information using an external network to perform scene relighting. All these methods have achieved remarkable performance in scene relighting. Nevertheless, those methods only handle daytime images for both input and output, neglecting the impact of internal light sources. As a result, they are inadequate for accomplishing effective day-to-night transformation. For day-to-night transformation, most works utilized generative methods, such as CycleGAN~\cite{CycleGAN2017}, pix2pix~\cite{pix2pix2017} and EnlightenGAN~\cite{jiang2021enlightengan}. Such purely data-driven approaches cannot accurately render spatially varying illumination, especially at night. Furthermore, although these methods sometimes do succeed in turning inactive light sources (e.g. street lights or windows) from off to on, the lights they produce are not accurate and realistic. Relighting daytime images to nighttime is also addressed in~\cite{Punnappurath_2022_CVPR}, which did not consider 3D geometry or materials and thus cannot model the interaction of light with the scene at night time. Moreover, nighttime-activated light sources are modelled in 2D instead of 3D, which leads to unrealistic illumination in the output image.

\section{Nighttime Driving Scene Understanding}
\label{sec:related_dataset}
Parsing and understanding the driving scene is a crucial ability for autonomous driving cars. Semantic segmentation has developed rapidly over the past few years and achieved remarkable progress. However, comprehension of nighttime driving scenes is still in its early stages, mainly due to the significant domain gap between daytime and nighttime scenes. Some works performed domain adaptation to close this gap.~\cite{Lengyel_2021_ICCV} Utilized a physics-based prior for domain adaptation, aiming to minimize the distribution shift between daytime and nighttime neural network feature maps.~\cite{2020_fda} then relied on the pixel-level adaptation via explicit transforms from source to target. An alternative method is to train traditional segmentation models on nighttime driving datasets, however, this requires annotated nighttime images which are hard to obtain. Though many datasets such as the Oxford RobotCar dataset and the BDD100K dataset have been including nighttime images~\cite{bdd100k, RCDRTKArXiv}, there has been a lack of emphasis on nighttime scene comprehension. As a result, these datasets do not offer adequate resources for training an effective model on nighttime image segmentation. A recently proposed autonomous driving dataset ACDC focused specifically on adverse conditions, contains 4006 images that are evenly distributed across four weather conditions: rain, fog, snow and night~\cite{SDV21ACDC}. Each image comes with a pixel-level semantic annotation and a reference image that is taken at the same location under normal conditions (clear daytime). Though the ACDC dataset puts a larger emphasis on nighttime (it includes 1006 nighttime images, with 400 from the training set, 106 from the validation set and 500 from the testing set), the gap still remains due to the shortage of annotated nighttime images caused by the difficulties of manual annotation.

\vspace{1cm}
\noindent Different from all methods discussed above, our method targets the generation of realistic nighttime images through simulation based on images from daytime datasets. In our image simulation pipeline, we utilize geometric information to reconstruct scene mesh and consider real-world light sources during relighting. As shown in the remaining sections of the paper, our work has the potential to close the gap in nighttime driving scene understanding.

%% file: text/3-materialsandmethods.tex
%
\newpage
\chapter{Materials and Methods}
\vspace{-7mm}
In this section, we first formalize the problem setting. Next, we introduce our data preparation procedure. Finally, we introduce our data generation pipeline involving geometry mesh reconstruction and realistic nighttime scene relighting.~\myfigref{fig:method} shows an overview of our data generation pipeline.

\section{Problem Setting}
\vspace{-2mm}
Our goal is to generate realistic nighttime images based on daytime images in the ACDC dataset~\cite{SDV21ACDC} reference split using a physics-based method. As described in~\mysecref{sec:related_dataset}, the ACDC dataset reference split contains 2006 daytime images from the training and validation reference set, in which 1003 of them come with a corresponding semantic annotation. Given those 1003 images and their semantic annotations, we propose a data processing pipeline that generates realistic nighttime images for each input image.
\section{Data Preparation}
\begin{figure}
    \centering
    \includegraphics[width=\textwidth]{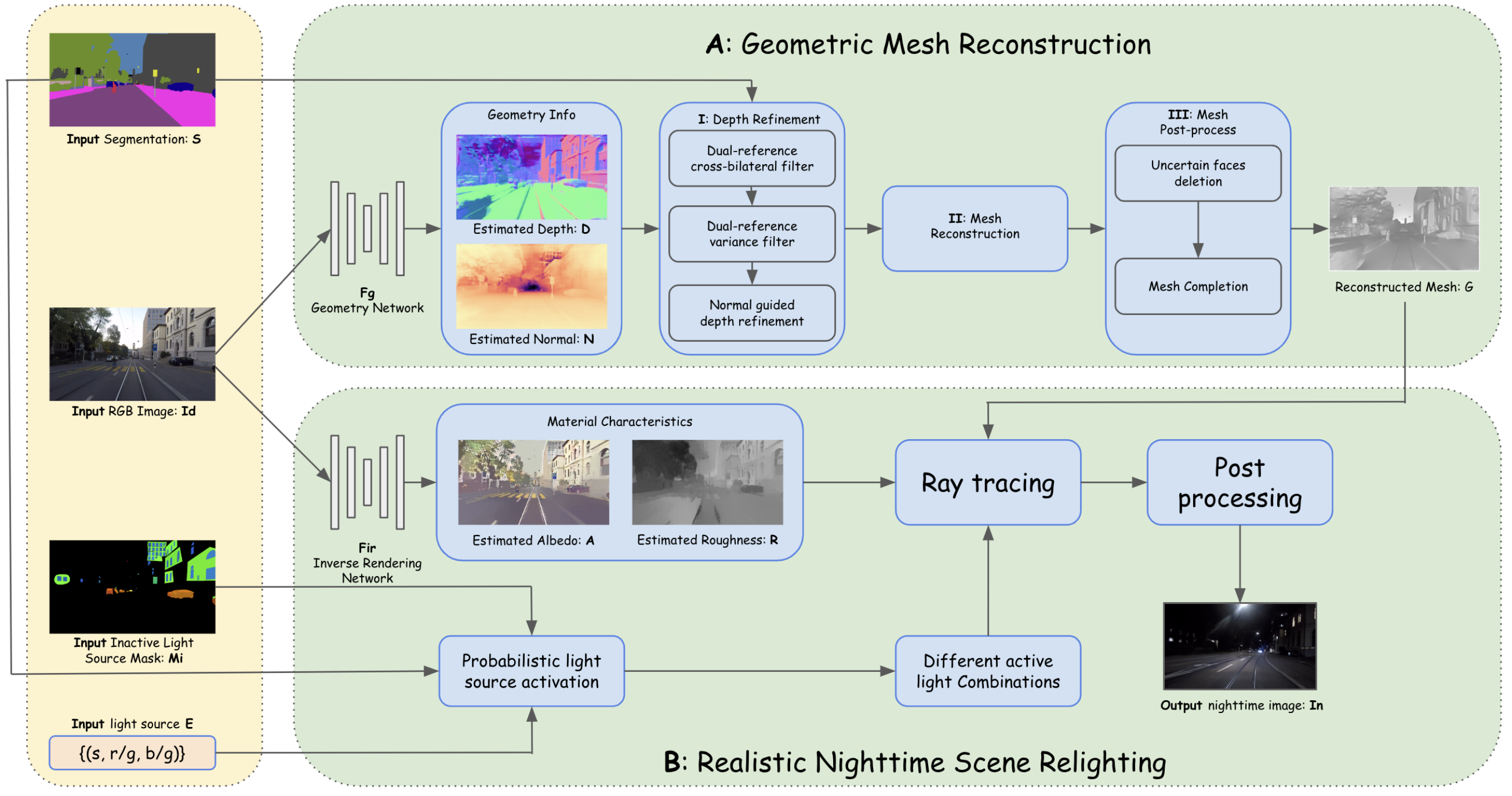}
    \vspace{-8mm}
    \caption{\textbf{Method overview.} Our pipeline contains two components. The \textbf{Geometric Mesh Reconstruction} component first utilizes network $\mathbf{F_g}$ to estimate the geometric information of an input RGB image, then reconstruct scene mesh based on depth using the Worldsheet~\cite{hu2021worldsheet} (\textbf{A.II}:~\mysecref{sec:gmr}). It also contains a depth refinement kernel (\textbf{A.I}:~\mysecref{sec:drk}) and a mesh post-process kernel (\textbf{A.III}:~\mysecref{sec:mpk}) to optimize depth and mesh, respectively. The \textbf{Realistic Nighttime Scene Relighting} (\textbf{B}:~\mysecref{sec:rnsr}) component first generates nighttime light sources using probabilistic light source activation. Then predict the material characteristics using network $\mathbf{F_{ir}}$. Following that, it uses ray tracing to render the linear nighttime clear image. Last, it processes the linear nighttime image to simulate artifacts and finally generates the output nighttime image $\mathbf{I_n}$. In this thesis, we implemented the Geometric Mesh Reconstruction component.}
    \label{fig:method}
\end{figure}

\vspace{-2mm}
\label{sec:data_prep}
 As shown in~\myfigref{fig:method}, except for daytime RGB images $I_d$ and its corresponding segmentation annotations $S$, our pipeline also takes a binary mask $M_i$ indicating the inactive light sources in $I_d$, and a set of nighttime light sources $E$ as input. In this section, we show the techniques employed for the creation of binary masks and the light sources.
 
\begin{figure}
\centering
\begin{tabular}{@{}c@{\hspace{1mm}}c@{}}
    \includegraphics[width=0.49\linewidth]{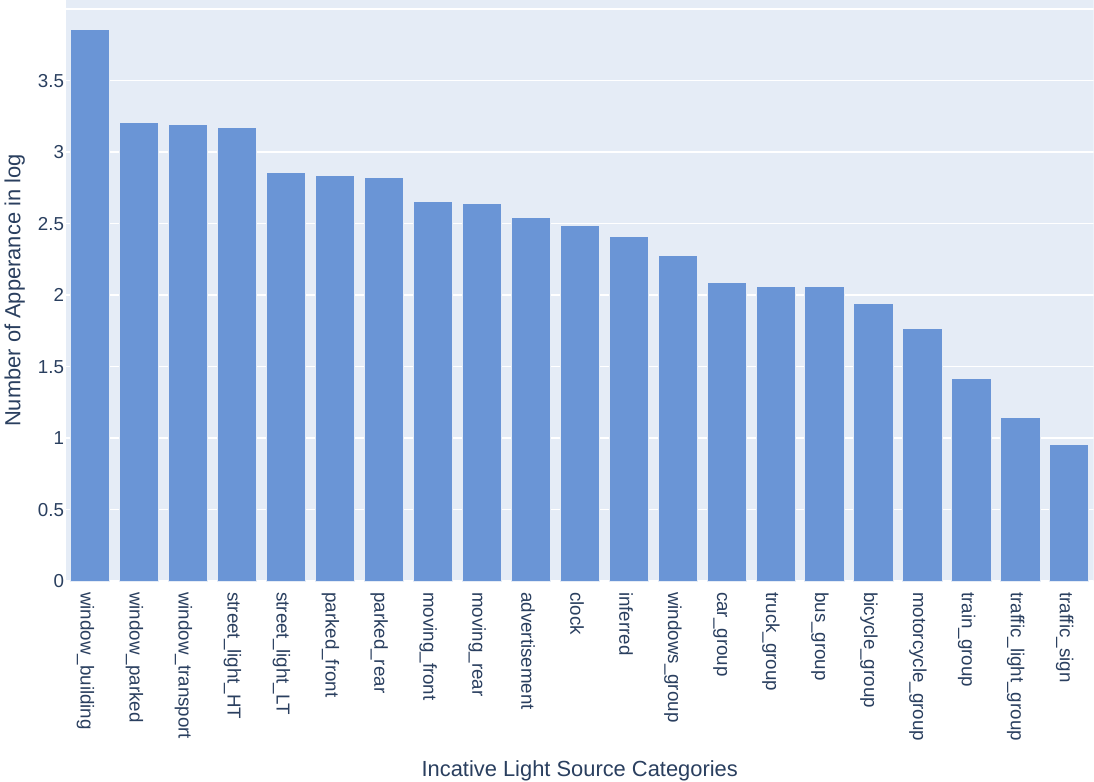} &
    \includegraphics[width=0.49\linewidth]{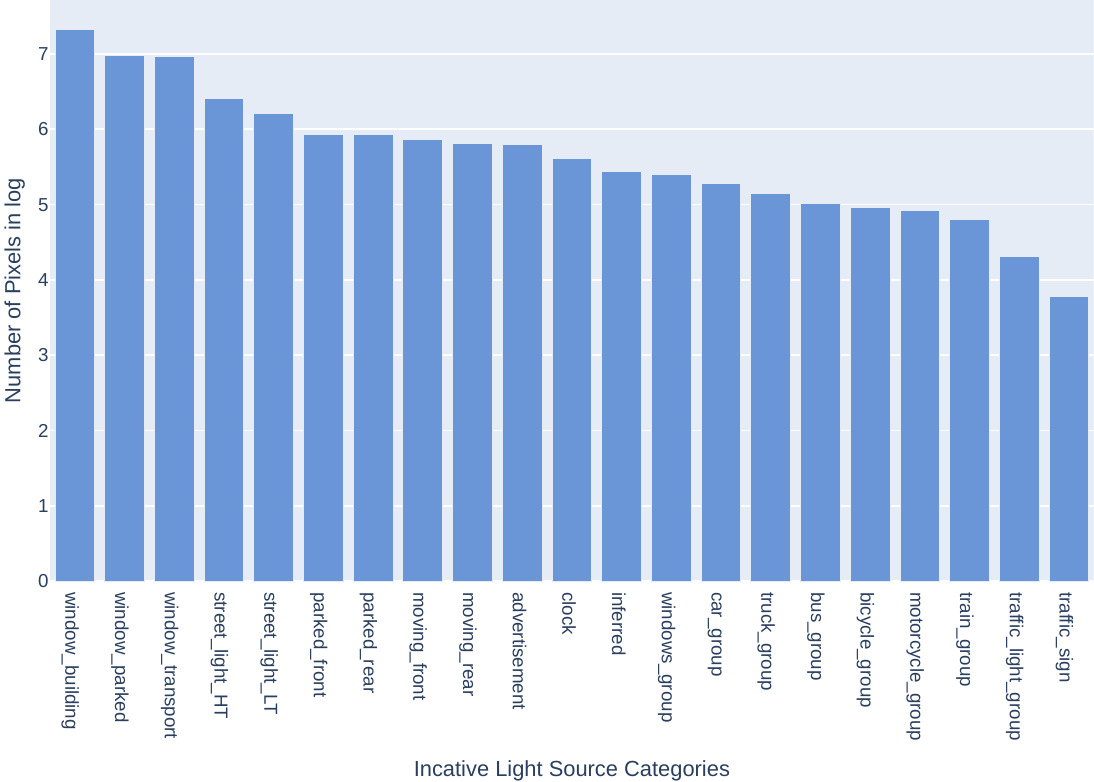}  \\
    \small (a) Instance Statistics & \small (b) Pixel Statistics \\
\end{tabular}
\caption{\textbf{Annotation Statistics.} We show the annotation statistics of 230 images nighttime reference images. The left image shows the number of instances of each type of light source, and the right image shows the number of pixels occupied by each type of light source. All results are presented in the base-10 logarithm.}
\label{fig:annot_stat}
\end{figure}

 \mypara{Binary mask $\mathbf{M_i}$} Though some works have already been focusing on light source separation~\cite{Yoshida_2023_CVPR}, they are not sufficient in such complex situations. Thus, we manually annotated inactive light sources using Segments.ai~\cite{Multi-sensor}, an online image labelling platform based on super-pixels. To control the quality of our annotations, two annotators first annotated 100 images, then conducted a cross-check by checking each other's annotations and then modifying the annotation rule until a consensus was reached. Ultimately, we defined 12 classes of inactive light sources and categorized them into three primary groups: buildings, vehicles, and objects. We also propose that light sources from the same category should be mutually related, for example, windows on the same floor or car lights from the vehicle. Thus, to ensure the interconnection between light sources, we also defined 9 group classes, see Table~\ref{tab:annot_class} for all classes and their definitions. In this thesis, we annotated 230 out of 253 images from the nighttime reference split (we filtered out 23 images as they were taken at twilight).~\myfigref{fig:annot_stat} shows the statistics of annotated pixels and the number of instances for each class.~\myfigref{fig:annot_example} shows some visual examples of our annotation. More examples can be found in~\myappendixref{apd:A}.
 
\mypara{Inactive light source $\mathbf{I_d}$} To make our nighttime images as realistic as possible, we planned to collect the light source dataset $E$ from a real-world setting. Following the method described in~\cite{Punnappurath_2022_CVPR}, we will first place gray cards under different nighttime illuminations (i.e. near different light sources identified in Table~\ref{tab:annot_class}). Next, we will capture images of each gray card to extract both chromaticity value $[\frac{r}{g}, \frac{b}{g}]$ and strength value $s$.
\begin{samepage}
\begin{table*}
\newcolumntype{Z}{S[table-format=2.3,table-auto-round]}
\centering
\vspace{-0.5em}
\setlength{\tabcolsep}{3mm}
\small
\footnotesize
\centering
\begin{tabular}{|c|c|c|c|}\hline
           No. & Name & Category & Detailed explanation \\\hline
           1   & window\_building  & building& Building windows that may emit light at night \\\hline
           2   & window\_parked   &  vehicle & Windows of parked vehicles, mainly cars\\\hline
           3   & parked\_front    &  vehicle & Front light of parked vehicles, mainly cars \\\hline
           4   & parked\_rear     &  vehicle & Rear light of parked cars, mainly cars \\\hline
           5   & moving\_front    &  vehicle & Front light of moving vehicles, mainly cars \\\hline
           6   & moving\_rear     &  vehicle & Rear light of moving vehicles, mainly cars \\\hline
           7   & window\_transport&  vehicle & Windows of public transportation that may emit light\\\hline
           8   & street\_light\_HT &  object & High temperature traffic lights, usually brighter \\\hline
           9   & street\_light\_LT &  object  & Low temperature traffic lights, usually dimer \\\hline
           10   & advertisement   &  object  & Advertisements that may emit light at night \\\hline
           11   & clock           &  object  & Clocks that emit light at night, mostly appear at bus stop \\\hline
           12  & inferred        &  object  & Light sources whose light colour can be inferred from its daytime color\\\hline
           13  & windows\_group   &  group   & Group of windows that belong to the same floor of the same building\\\hline
           14  & car\_group       &  group   & Group of light sources that belong to the same car\\\hline
           15  & truck\_group     &  group   & Group of light sources that belong to the same truck\\\hline
           16  & bus\_group       &  group   & Group of light sources that belong to the same bus\\\hline
           17  & bicycle\_group   &  group   & Group of light sources that belong to the same bicycle\\\hline
           18  & motorcycle\_group&  group   & Group of light sources that belong to the same motorcycle\\\hline
           19  & train\_group     &  group   & Group of light sources that belong to the same train     \\\hline
           20  & traffic\_light\_group& group & Group of traffic lights that belong to the same panel\\\hline
           21  & traffic\_sign\_group & group & Group of light sources that belong to the same sign \\\hline
\end{tabular}
\vspace{-2mm}
\caption{{\bf Inactive light source class and their definition}. We defined 21 types of inactive light sources, each of them belonging to one category, including building, vehicle, object and group.}
\label{tab:annot_class}
\end{table*}
\begin{figure}
\centering
\begin{tabular}{@{}c@{\hspace{1mm}}c@{\hspace{1mm}}c@{}}
\includegraphics[width=0.325\linewidth]{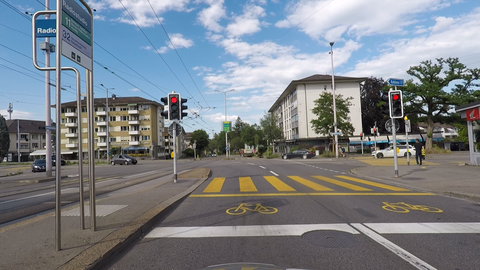} &
\includegraphics[width=0.325\linewidth]{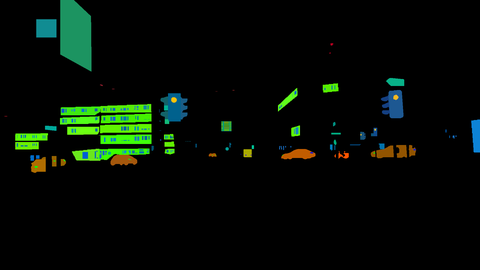}  &
\includegraphics[width=0.325\linewidth]{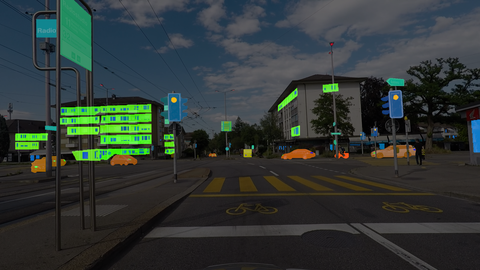}\\
\includegraphics[width=0.325\linewidth]{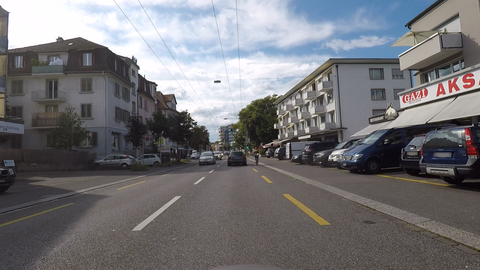} &
\includegraphics[width=0.325\linewidth]{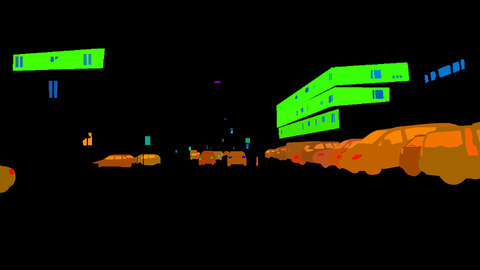}  &
\includegraphics[width=0.325\linewidth]{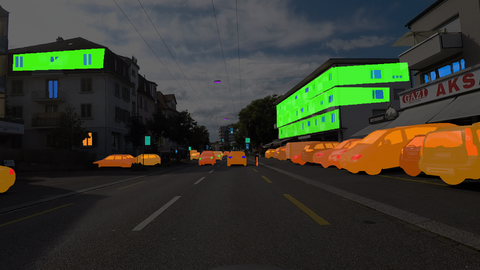}\\
\includegraphics[width=0.325\linewidth]{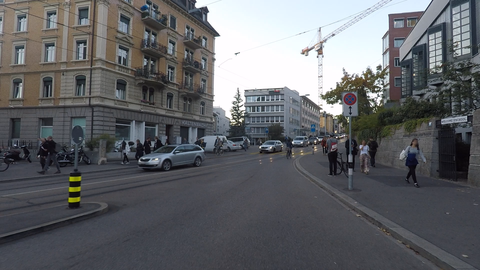} &
\includegraphics[width=0.325\linewidth]{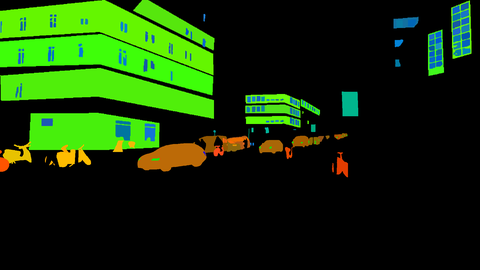}  &
\includegraphics[width=0.325\linewidth]{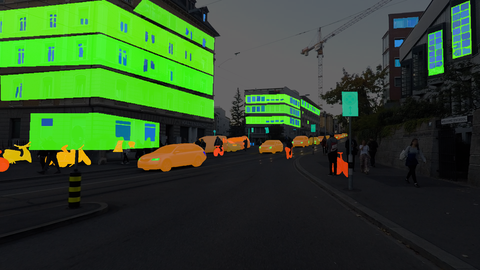}\\
\small (a) Input daytime RGB image & \small (b) Inactive light source mask & 
\small (c) Superposition of (a) and (b)\\
\end{tabular}
\vspace{-2mm}
\caption{\textbf{Annotation Examples.} We present several examples of our inactive light source annotation. The left column shows the input daytime RGB image, the middle column shows the annotated inactive light source mask, where each instance has its own identity and bounding box, and the right column superimposes the RGB image and the light source mask.}
\label{fig:annot_example}
\end{figure}
\end{samepage}
\section{Geometric Mesh Reconstruction}
\label{sec:gmr}
The first component of our pipeline involves reconstructing scene mesh from a daytime RGB image and its corresponding segmentation annotation. Depth and surface normal are key geometric information that we rely on to reconstruct the mesh. In particular, we use iDisc~\cite{piccinelli2023idisc} as out Geometry Network ($F_g$) to estimate depth ($D$) and normal ($N$). For depth estimation, we directly utilize the model pre-trained on the KITTI dataset~\cite{Geiger2012CVPR}. For normal estimation, the pre-trained model on the NYUv2 dataset~\cite{Silberman:ECCV12} doesn't adapt well to the ACDC dataset~\cite{SDV21ACDC}, so we retrain the iDisc model using the DIODE dataset~\cite{diode_dataset} outdoor split, initializing the training with pre-trained swin transformer~\cite{liu2021Swin} weight. 

Inspired by previous work SIMBAR~\cite{zhang2022simbar}, we apply Worldsheet~\cite{hu2021worldsheet}, a novel view geometry scene synthesis method to reconstruct scene mesh. Worldsheet builds a scene mesh by warping a grid sheet onto the scene geometry via grid offset and depth. The grid exhibits a horizontal offset $\Delta \hat{x}$ and a vertical offset $\Delta \hat{y}$. Importantly, there is no need for predicting or adjusting individual vertices in the $x$ and $y$ directions. The mesh vertices are formulated from grid offset and depth as:
\begin{equation}
    V_{w, h}=\left[\begin{array}{c}
    d_{w, h} \cdot\left(\hat{x}_{w, h}+\Delta \hat{x}_{w, h}\right) \cdot \tan \left(\theta_F / 2\right) \\
    d_{w, h} \cdot\left(\hat{y}_{w, h}+\Delta \hat{y}_{w, h}\right) \cdot \tan \left(\theta_F / 2\right) \\
    d_{w, h}
\end{array}\right]
\end{equation}
Where $d$ denotes the external depth, $x$ and $y$ denote the horizontal and vertical location of vertices in the mesh coordinates equally spaced from $ - 1$ to $1$, and $\theta_F$ is the camera field of view. Thus, we are able to reconstruct the scene mesh based on external depth. However, this requires that the input depth maps we provide have high precision. In the next two sections, we will introduce the Depth Refinement Kernel that optimizes depth before the reconstruction and the Mesh Post-process Kernel after the reconstruction.

\section{Depth Refinement Kernel}
\label{sec:drk}
The main purpose of the Depth Refinement Kernel is to optimize depth based on segmentation annotations and predicted normal map, where we treat predicted normal as ground truth during optimization. This kernel consists of three parts: Dual-reference cross-bilateral filter, Dual-reference variance filter and Normal guided depth refinement.

\subsection{Dual-reference Cross-bilateral Filter}
\begin{figure}
\centering
\begin{tabular}{@{}c@{\hspace{1mm}}c@{\hspace{1mm}}c@{}}
\includegraphics[width=0.325\linewidth]{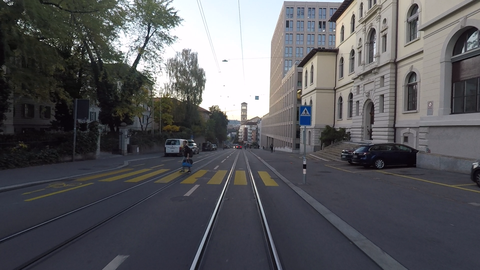} &
\includegraphics[width=0.325\linewidth]{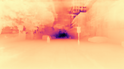}  &
\includegraphics[width=0.325\linewidth]{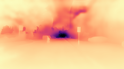}\\
\includegraphics[width=0.325\linewidth]{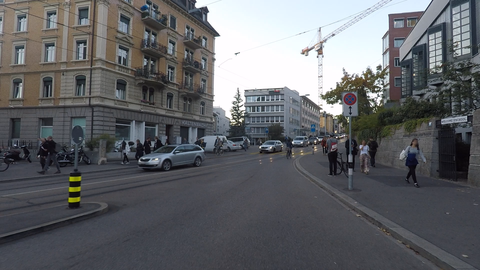} &
\includegraphics[width=0.325\linewidth]{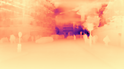}  &
\includegraphics[width=0.325\linewidth]{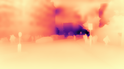}\\
\small (a) Original RGB image & \small (b) Depth before & 
\small (c) Depth after\\
\end{tabular}
    \caption{\textbf{Depth comparison before and after the Dual-reference cross-bilateral filter.} We present two examples for depth comparison. Column \textbf{(a)} shows the original RGB image, column \textbf{(b)} shows the depth estimated by iDisc~\cite{piccinelli2023idisc}, and column \textbf{(c)} shows the depth after dual-reference cross-bilateral filter optimization. The above comparison shows that the dual-reference cross-bilateral filter improves the depth estimation at the pixel level.}
    \vspace{-2mm}
    \label{fig:depth_comparison}
\end{figure}

Inspired by~\cite{SDHV18}, we apply the Dual-reference cross-bilateral filter as the first building block of our Depth Refinement Kernel. In our work, we optimize input depth $\hat{d}$ using the RGB image and the semantic annotation to obtain filtered depth $d$, shown as~\myeqref{equ:cross-bilateral}.
\begin{equation}
\label{equ:cross-bilateral}
d(\mathbf{p})=\frac{\sum_{q \in \mathcal{N}(\mathbf{p})} G_{\sigma_s}(\|\mathbf{q}-\mathbf{p}\|)\left[\delta(h(\mathbf{q})-h(\mathbf{p}))+\mu G_{\sigma_c}(\|\mathbf{J}(\mathbf{q})-\mathbf{J}(\mathbf{p})\|)\right] \hat{d}(\mathbf{q})}{\sum_{q \in \mathcal{N}(\mathbf{p})} G_{\sigma_s}(\|\mathbf{q}-\mathbf{p}\|)\left[\delta(h(\mathbf{q})-h(\mathbf{p}))+\mu G_{\sigma_c}(\|\mathbf{J}(\mathbf{q})-\mathbf{J}(\mathbf{p})\|)\right]}
\end{equation}
Similar to~\cite{SDHV18}, we use the CIELAB counterpart of the input RGB image $\mathcal{R}$, denoted by $\mathbf{J}$. Where $\mathbf{p}$, $\mathbf{q}$ means pixel locations, $\mathcal{N}$ means neighbouring pixels and $h$ means semantic classes. $\delta$ is the Kronecker delta, G denotes the Gaussian kernel, where $G_{\sigma_s}$ is the spatial Gaussian kernel and $G_{\sigma_c}$ is the colour Gaussian kernel lead by constant $\mu$. The numerator and denominator of this equation consist of two main components, in which $\delta(h(\mathbf{q})-h(\mathbf{p}))$ is for semantic references and  $\mu G_{\sigma_c}(\|\mathbf{J}(\mathbf{q})-\mathbf{J}(\mathbf{p})\|)$ is for colour references. The semantic component of this equation infers that only pixels from different semantic classes can contribute to this term, making the edge of each semantic object sharper. At the same time, the colour component helps to preserve depth that can be implied from the colour variation of the input RGB image. Following~\cite{SDHV18}, we initially set $\mu=5$ and $\sigma_s=10$. For the colour component, we discovered that when its weight grows larger, the depth change at semantic edges will become smoother and more spurious faces will be created (as described in~\mysecref{sec:variance_filter}). Thus, we decreased its weight and set $\sigma_c=5$.~\myfigref{fig:depth_comparison} shows the comparison of depth maps before and after the Dual-reference cross-bilateral filter.
\subsection{Dual-reference Variance Filter}
\label{sec:variance_filter}
\begin{figure}
\centering
\begin{tabular}{@{}c@{\hspace{1mm}}c@{\hspace{1mm}}c@{}}
\includegraphics[width=0.325\linewidth]{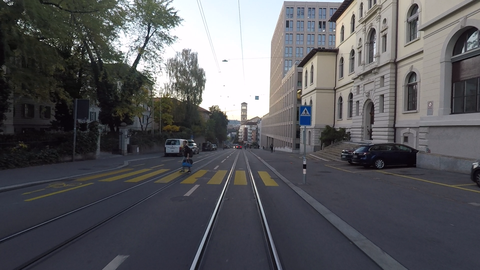} &
\includegraphics[width=0.325\linewidth]{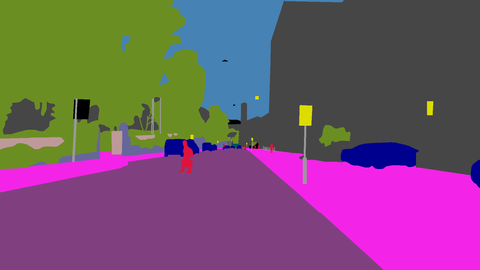}  &
\includegraphics[width=0.325\linewidth]{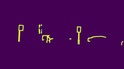}\\
\includegraphics[width=0.325\linewidth]{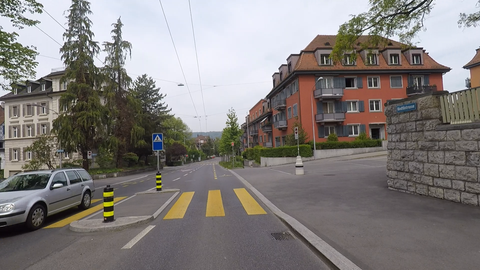} &
\includegraphics[width=0.325\linewidth]{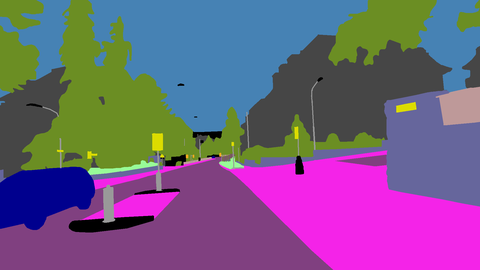}  &
\includegraphics[width=0.325\linewidth]{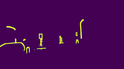}\\
\small (a) Original RGB image & \small (b) Semantic annotation & 
\small (c) Uncertain map\\
\end{tabular}
\caption{\textbf{Uncertain region detected by the Dual-reference variance filter.} Column \textbf{(a)} shows the original RGB image, column \textbf{(b)} shows the semantic annotation and column \textbf{(c)} shows the generated uncertain map, where the uncertain region is represented by the yellow region.}
\vspace{-2mm}
\label{fig:uncertain_region}
\end{figure}

One disadvantage of the Worldsheet~\cite{hu2021worldsheet} is that its generated scene mesh is an equal offsets mesh grid with only one layer of vertices. This creates spurious faces at depth discontinuities, connecting foreground objects with background objects. Although those spurious faces are not visible from the camera angle, they will still generate unrealistic reflections during the final relighting process. To solve this issue, we designed a Dual-reference variance filter to identify spurious faces based on depth maps and semantic annotations, as shown in~\myeqref{equ:cross-variance}. We propose that spurious faces usually happen at uncertain regions that meet the following two criteria: \textbf{(1)} regions that contain at least one semantic boundary. \textbf{(2)} regions that have a large variation in depth. For instance, consider a scenario where a region includes both a section of a moving car and a segment of the road or perhaps a part of a traffic sign alongside a portion of a building. We further define a set of foreground objects that are used to identify semantic boundaries, including vehicles, persons, poles, traffic lights and traffic signs.
\begin{equation}
\label{equ:cross-variance}
\mathcal{U}(r(\mathbf{p}, l)) = (\mathcal{V}(d(r(\mathbf{p}, l))) > \mu) \text{ and } (\mathcal{V}(h(r(\mathbf{p}, l))) > 0)
\end{equation}
In~\myeqref{equ:cross-variance}, $U(\cdot)$ denotes the binary value of the uncertain map, $\mathbf{p}$ denotes pixel location and $r(\mathbf{p}, l)$ represents the square region with a size of $l * l$ pixels with $\mathbf{p}$ as the upper left corner. The two components of~\myeqref{equ:cross-variance} correspond to depth and semantic annotations, respectively. In the depth component $d(\cdot)$, we alert the region to be uncertain if the variance of all depth inside region $r(\mathbf{p}, l)$ is larger than a constant $\mu$. In the semantic component $h(\cdot)$, we alert the region if variance of all semantic value inside region $r(\mathbf{p}, l)$ is larger than zero: when a semantic change happens. Two components are connected by a logic and operator, meaning the region will be marked as uncertain if both components are alerted simultaneously. In our experiment, we set $\mu = 0.001$ and $l = 8$.~\myfigref{fig:uncertain_region} shows two visual examples of the uncertain region detected by the Dual-reference variance filter.
\subsection{Normal-Guided Depth Refinement}
Differentiable optimization has proved useful to refine and increase the accuracy of learning-based method results~\cite{turpin2023fastgraspd, zhang2023handypriors}. In our reconstruction problem, depth estimation provides coarse geometric information regarding the input image, while surface normal offers further intricate local details. To increase depth accuracy, we propose an optimization-based depth refinement method based on surface normal. For each input image, we formulate a loss function based on the interrelationship between its depth and surface normal, then use gradient descent to optimize depth by minimizing the loss. Next, we will describe the loss terms we used for the normal-guided depth refinement.

\mypara{Normal loss.} Given a depth map, we can infer the surface normal by computing the cross product of gradient vectors between neighbouring pixels, as shown in~\myeqref{equ:compute_normal}.
\begin{equation}
\label{equ:compute_normal}
\hat{\vec{N}} = \vec{\nabla{X}} \times \vec{\nabla{Y}} = (1, 0, \frac{\partial z}{\partial x}) \times (0, 1, \frac{\partial z}{\partial y}) = (-\frac{\partial z}{\partial x}, -\frac{\partial z}{\partial y}, 1)
\end{equation}
Where $\frac{\partial z}{\partial x}$ and $\frac{\partial z}{\partial y}$ denotes the gradient of depth with respect to $x$ and $y$ in the camera space, which can be computed via chain rule $\frac{\partial z}{\partial x} = \frac{\partial z}{\partial u} \cdot \frac{\partial u}{\partial x} $ and $\frac{\partial z}{\partial y} = \frac{\partial z}{\partial v} \cdot \frac{\partial v}{\partial y}$. And the transformation between pixel space and camera space is shown as~\myeqref{equ:transform_x_y}.
\begin{align}
\label{equ:transform_x_y}
u \cdot d = f_x \cdot x + c_x \longleftrightarrow \frac{\partial u}{\partial x} = \frac{f_x}{d} \\ \nonumber
v \cdot d = f_y \cdot y + c_y \longleftrightarrow \frac{\partial v}{\partial y} = \frac{f_y}{d}
\end{align}
Where $f_x$ and $f_y$ denote the focal length along the $x$ and $y$ axis, $c_x$ and $c_y$ denote the principal point along the $x$ and $y$ axis, respectively. The final expression of this loss is formalized in~\myeqref{equ:normal_loss}. In this equation, we treated depth estimated by iDisc~\cite{piccinelli2023idisc} $\vec{N_{\text{est}}}$ as ground truth.
\begin{equation}
\label{equ:normal_loss}
L_{\text{normal}} = \lVert \vec{\hat{N}} - \vec{N_{\text{est}}} \rVert_{2}^{2}
\end{equation}

\mypara{Continuity loss.} Although the normal loss can optimize depth to align its inferred normal with the reference surface normal, it lacks the ability to account for sudden variations in depth. As a result, this creates spurious faces, as discussed in Section~\ref{sec:variance_filter}. To tackle this issue, we proposed a continuity loss that directly optimizes gradient vectors based on surface normal, formalized as~\myeqref{equ:continuity_loss}.
\begin{equation}
\label{equ:continuity_loss}
L_{continuity} = \frac{1}{n} \sum_{i=1}^{n} ((\vec{\nabla X_\text{i}} \cdot \vec{N_\text{i}})^2 + (\vec{\nabla Y_\text{i}} \cdot \vec{N_\text{i}})^2) \cdot (1 - \mathcal{U_\text{i}})
\end{equation}
In this equation, $\mathcal{U}$ is the uncertain map derived by the dual-reference variance filter in~\myeqref{equ:cross-variance}, $\vec{\nabla X}$ and $\vec{\nabla Y}$ are gradient vectors computed in ~\myeqref{equ:compute_normal}, $n$ is the total number of pixels of the input image and $i$ represents the index of currently computing pixel. This loss term is aware of depth discontinuities as the sudden depth change between neighbouring pixels will create a gradient vector that is opposite from the normal vector, resulting large value of the dot product. Thus, it can help eliminate spurious faces created by the normal loss. The participation of the uncertain mask term $(1 - \mathcal{U_\text{i}})$ deals with the case when foreground objects have similar normal as background objects, avoiding the optimization process pushing them into the background.

\mypara{Depth loss.} The optimization process should respect the initial depth predicted by iDisc~\cite{piccinelli2023idisc}, meaning that the optimized depth should not deviate significantly from the initial estimation. Thus, we add a depth loss that punishes any depth change with respect to the estimated depth, shown as~\myeqref{equ:depth_loss}.
\begin{equation}
\label{equ:depth_loss}
L_{depth} = \lVert \hat{d} - d_{est} \rVert_{2}^{2}
\end{equation}

In summary, the final loss we used to optimize depth is the weighted sum of each individual loss, shown in~\myeqref{equ:final_loss}, where $\lambda$’s are the weights for the loss terms. 
\begin{equation}
\label{equ:final_loss}
L_{final} = \lambda_1 L_{normal} + \lambda_2 L_{continuity} + \lambda_3 L_{depth}
\end{equation}
In our experiments, we applied grid search to determine those hyper-parameters and set $\lambda_1 = 1$, $\lambda_2 = 1$ and $\lambda_3 = 5$, more details can be found in~\myappendixref{apd:B}. With those differentiable loss terms, we optimize the predicted depth using the gradient-based optimizer Adam~\cite{Adam2015} for $1000$ steps with a learning rate of $0.0001$.

\section{Mesh Post-processing Kernel}
\label{sec:mpk}

As explained in Section~\ref{sec:variance_filter}, the scene mesh reconstructed by Worldsheet~\cite{hu2021worldsheet} constitutes a single-layer mesh grid based on conventional depth map prediction. Consequently, its outcomes contain spurious faces and inaccurately link foreground and background objects within the invisible areas of the input image. To solve this issue,~\cite{wimbauer2023behind} proposed to predict the density field of the input image and map the location in the frustum to volumetric density in order to learn the real 3D feature. However, the performance of their method on complex scenes such as images in the ACDC~\cite{SDV21ACDC} dataset is limited. To bridge the gap, we designed a Mesh Post-process Kernel that contains two steps: uncertain faces deletion and mesh completion.
\subsection{Uncertain Faces Deletion}
The first part of Mesh Post-process Kernel deals with the removal of spurious faces created by the Worldsheet~\cite{hu2021worldsheet}. Specifically, we first perform re-projection of all vertices from 3D to 2D and flagged vertices located within the uncertain area, as described in Section~\ref{sec:variance_filter}, as uncertain vertices. Furthermore, we defined faces that contain at least one uncertain vertex as spurious faces. Following this, we traverse through all faces and remove those that are identified as spurious faces. The resulting object contains the foreground and background object mesh separately, as well as point clouds between them.

\subsection{Mesh Completion}
\begin{figure}
\centering
\begin{tabular}{@{}c@{\hspace{1mm}}c@{\hspace{1mm}}c@{}}
\includegraphics[width=0.325\linewidth]{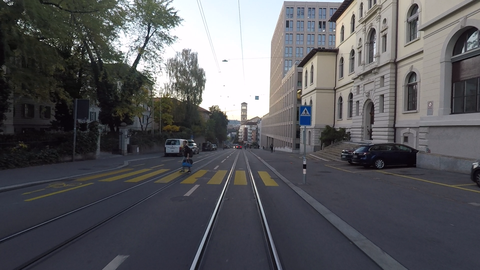}  &
\includegraphics[width=0.325\linewidth]{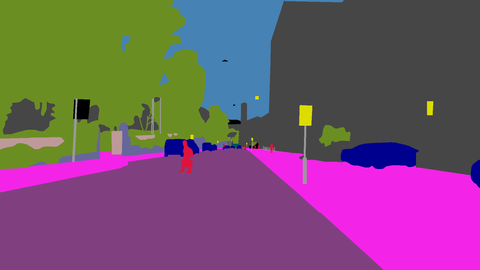}  &
\includegraphics[width=0.325\linewidth]{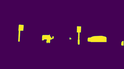}  \\
\includegraphics[width=0.325\linewidth]{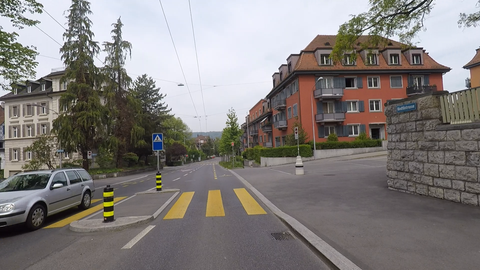}  &
\includegraphics[width=0.325\linewidth]{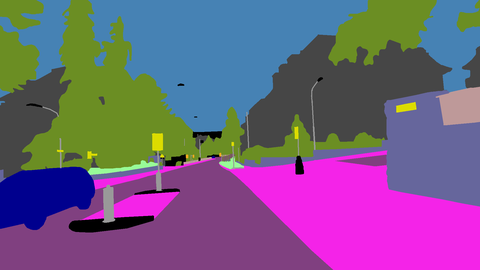}  &
\includegraphics[width=0.325\linewidth]{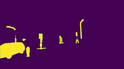}  \\
\small (a) RGB input & \small (b) Semantic annotation &  \small (c) New uncertain region \\
\end{tabular}
\caption{\textbf{Completed uncertain region.} Column (\textbf{a}) shows the original RGB image, column (\textbf{b}) shows the semantic annotation and column (\textbf{c}) shows the completed uncertain region, where the uncertain region is represented by the yellow region.}
    \label{fig:mesh_complete}
\end{figure}

The first part of the Mesh Post-process Kernel introduces holes in the mesh surface. The second part is designed to complete these holes and ensure the mesh surface becomes watertight. To achieve this, we first determine regions that need to be completed by finding the union of uncertain regions and their neighbouring foreground semantic segments. After that, we complete the mesh sheet by adding vertices into the newly determined uncertain region. In particular, for each newly determined uncertain region, we break it into horizontal lines with one-pixel height. Then adding linearly distributed vertices based on their left and right vertices location. In the end, we add faces between newly added vertices and boundary vertices to make the mesh sheet watertight.~\myfigref{fig:mesh_complete} shows an example of the newly determined uncertain region, and Algorithm 1 shows the operation process. Furthermore, to ensure the pixel-level accuracy of foreground objects, we also apply mesh completion to foreground objects. For the intersection of uncertain regions and foreground object semantics, we select a pixel that is adjacent to the persistent foreground semantics. Here, we assigned the vertex depth as the average of its neighbouring vertices. Following that, we extend this operation to all pixels within the intersection region by employing a breadth-first search approach. Lastly, we establish connections between the newly introduced foreground vertices and their corresponding persisting foreground mesh, effectively rendering each foreground object watertight.
\begin{algorithm}
\label{alg:mesh}
\caption{Background Mesh Completion Algorithm}\label{alg:cap}
\begin{algorithmic}[1]
\Require Binary mask list $M$ indicating new uncertain maps, Incomplete mesh sheet $G_u$
\Ensure Each mask maps to a region that needs to be completed on the mesh sheet
\For {Every mask $m \in M$}
    \State $I_t \gets$ top row index of $m$ that has missing vertex
    \State $I_b \gets$ bottom row index of $m$ that has missing vertex
    \For {$i = I_t$ to $I_b$}
        \State $J_l \gets$ index of the leftmost vertex row $i$ that is missing
        \State $J_r \gets$ index of the rightmost vertex row $i$ that is missing
        \State $G_l \gets G_u[i][J_l-1]$
        \State $G_r \gets G_u[i][J_r+1]$
        \For {$j = J_l$ to $J_r$}
            \State $G_u[i][j] \gets \frac{G_l(j_r-j+1)+G_r(j-j_l+1)}{j_r-j_l+2}$ \Comment{Add vertices uniformly between $G_l$ and $G_r$}
        \EndFor
    \EndFor
\EndFor
\State Connect vertices in $G_u$ to form faces \Comment{Make G watertight triangle mesh}
\end{algorithmic}
\end{algorithm}
\section{Realistic Nighttime Scene Relighting}
\label{sec:rnsr}
The second component of our pipeline aims to relight the reconstructed scene for realistic nighttime images based on material characteristics and nighttime light sources, which involves four steps. Though we did not implement this component in this thesis, we will explain the detailed method in the rest of this section. First of all, we utilize an inverse rendering network $\mathbf{F_{ir}}$ to predict the material characteristics, and then we will use a probabilistic light source activation to generate light sources. After that, we plan to apply the traditional rendering technique ray tracing to render clear linear nighttime images. In the end, we will also run post-processing to the clear nighttime image to simulate artifacts caused by the camera, including exposure time, noise and ISP. The remaining part of this section illustrates each step in detail.

\subsection{Material Characteristics Prediction}
Many prior studies have been focused on predicting material characteristics on small objects~\cite{Li_2023_CVPR, Jin_2023_CVPR} or indoor scenes~\cite{neuralSengupta19, Wu_2023_CVPR, Munkberg_2022_CVPR, li2020inverse}. Nevertheless, only a few of them have made an attempt to tackle the same task with outdoor datasets~\cite{Wang_2023_CVPR}. In our work, we explored different existing methods and tested their performance on a synthetic outdoor optical flow dataset MPI Sintel~\cite{Butler:ECCV:2012}. Our testing results show that most indoor methods demonstrate good generalization capabilities on the outdoor dataset, likely due to the relatively small disparity between outdoor and indoor materials. Ultimately, based on these findings, we employe~\cite{li2020inverse} as our inverse rendering network $\mathbf{F_{ir}}$ to predict albedo and roughness from the ACDC~\cite{SDV21ACDC} RGB images. In the future, we will also test the performance of~\cite{Wang_2023_CVPR} once it is released and adapt it into our pipeline if it is proved to be better than~\cite{li2020inverse}.

\subsection{Probabilistic Light Source Activation and Relighting}
\label{sec:light_active}
Unlike most scene relighting methods that consider sunlight as the only light source~\cite{lyu2022nrtf, srinivasan2021cvpr, Yang_2023_CVPR}, our day-to-night simulation pipeline considers over 30 inactive light sources per scene (shown in~\myfigref{fig:annot_stat}). Thus, to generate realistic and diverse nighttime images, we design the probabilistic light source activation kernel to manage the activation of each light source. We define the activation of each light source as an independent random variable with a Bernoulli Distribution $f(\mathbf{p}, \mathbf{l})$, where $\mathbf{p}$ is the probability for event $\textbf{l}$ to happen, meaning that there is a  probability of $p$ for the light source to be active and $(1-p)$ to be inactive. Furthermore, we also consider that light sources exhibit interdependence in the real world. For instance, adjacent windows are likely to be active or inactive simultaneously, front light and rear light of a moving car are also to be active together. To capture this real-life characteristic, we utilize group masks, as defined in \mysecref{sec:data_prep}, to group light sources together. This grouping approach ensures that each group of light sources shares the same $\mathbf{p}$ parameter yet remains independent from other groups. By incorporating this interdependency feature into our light source activation model, we can accurately emulate real-world light sources' behaviour and enhance our simulations' authenticity. Once the scene mesh, material attributes, and light sources are established, we perform ray tracing~\cite{1980raytracing} to render the nighttime image. For each daytime image, we employ various activation parameters for the inactive sources, creating multiple nighttime images.

\subsection{Image Post-processing}
Nighttime images usually contain artifacts such as noise caused by low illumination and long exposure time at night. To simulate this real-world scenario, we will apply post-processing to clear linear images generated by ray tracing. Following~\cite{Punnappurath_2022_CVPR}, we plan to adopt the well-established heteroscedastic Gaussian model~\cite{2009_noise, 2008_noise, 2014_noise, 2013_noise} for noise. Given a nighttime clear image $I_c$, we will generate the nighttime noisy image $I_n$ with the following equation:
\begin{equation}
    \mathbf{I}_n = \mathbf{I}_c+\mathcal{N}\left(\mathbf{0}, \beta_1 \mathbf{I}_{\text {c}}+\beta_2\right)
\end{equation}
Where $\beta_1$ and $\beta_2$ are shot and read noise parameters, which we empirically determined based on measuring the noise of real noisy/clean nighttime image pairs for different ISO levels.

%% file: text/4-experimentsandresults.tex
%
\newpage
\chapter{Results and Experiment Plans}
\vspace{-3mm}



In this section, we will apply our pipeline to images within the ACDC dataset and present the results of the Geometry Mesh Reconstruction component. Additionally, we will demonstrate the generalizability of our pipeline by applying it to other autonomous driving datasets. such as the Cityscapes dataset~\cite{Cordts2016Cityscapes}. Furthermore, we will outline our testing plans for the entire pipeline.
\vspace{-2mm}
\section{Geometry Mesh Reconstruction}
\subsection{Datasets and Metrics}
\label{sec:dataset}
\mypara{Datasets} In this work, we used images in the ACDC dataset~\cite{SDV21ACDC}, which is a large-scale dataset consisting of 4006 images evenly distributed across four different adverse conditions: snow, fog, rain and night. Each adverse condition image comes with a high-quality fine pixel-level semantic annotation and a reference image of the same scene taken under normal conditions (clear daytime). Our work mainly focused on applying our pipeline to the reference split of the ACDC dataset. Except for the ACDC dataset, we also use other autonomous driving datasets, including the BDD100K dataset~\cite{bdd100k}, Cityscapes dataset~\cite{Cordts2016Cityscapes} and the Dark Zurich dataset~\cite{SDV19}. In particular, we use daytime images and its corresponding semantic annotations for all selected datasets, showing that our pipeline has a strong adaptability.

\mypara{Metrics} The Geometry Mesh Reconstruction component generates scene mesh and geometric information such as depth and surface normal. However, none of the datasets mentioned above provide ground truth for evaluation. Thus, we conduct a qualitative comparison of the generated mesh and its surface normals to evaluate the Geometry Mesh Reconstruction component of our pipeline.

\subsection{Mesh Comparison}
\begin{figure}
\label{fig:mesh_comp}
\centering
\begin{tabular}{@{}c@{\hspace{1mm}}c@{\hspace{1mm}}c@{\hspace{1mm}}c@{\hspace{1mm}}c@{\hspace{1mm}}c@{}}
\includegraphics[width=0.19\linewidth]{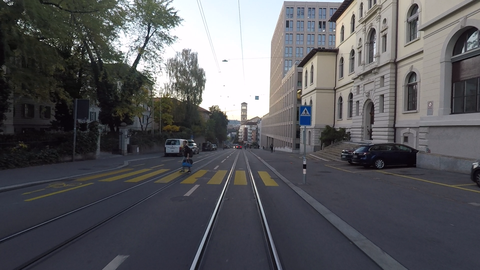}  &
\includegraphics[width=0.19\linewidth]{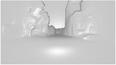}  &
\includegraphics[width=0.19\linewidth]{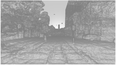}  &
\includegraphics[width=0.19\linewidth]{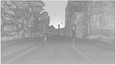}  &
\includegraphics[width=0.19\linewidth]{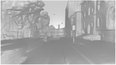}  \\
\includegraphics[width=0.19\linewidth]{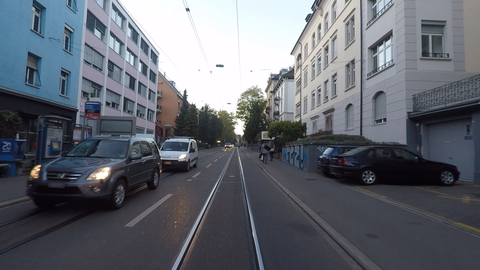}  &
\includegraphics[width=0.19\linewidth]{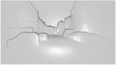}  &
\includegraphics[width=0.19\linewidth]{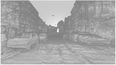}  &
\includegraphics[width=0.19\linewidth]{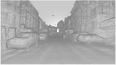}  &
\includegraphics[width=0.19\linewidth]{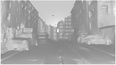}  \\
\small (a) RGB input & \small (b) (1) &  \small (c) (2) & \small (c) (3) & \small (d) Ours\\
\end{tabular}
    \caption{\textbf{Mesh reconstruction result comparison.} We compared the mesh reconstruction result with mesh reconstructed by other methods with the following settings, (1): Worldsheet with MiDaS depth, (2): Worldsheet with iDisc depth, (3): SIMBAR reconstruction. Our method can preserve more accurate geometric information and construct smoother mesh surfaces than all other methods.}
    \label{fig:mesh_comparison}
\end{figure}

\begin{figure}
\centering
\begin{tabular}{@{}c@{\hspace{1mm}}c@{\hspace{1mm}}c@{\hspace{1mm}}c@{\hspace{1mm}}c@{\hspace{1mm}}c@{\hspace{1mm}}c@{}}
\includegraphics[width=0.155\linewidth]{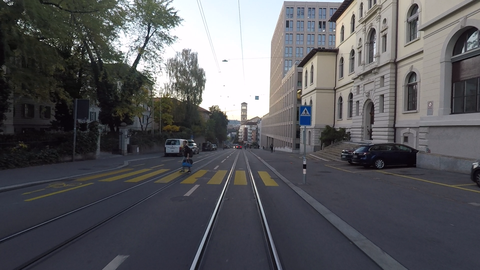}  &
\includegraphics[width=0.155\linewidth]{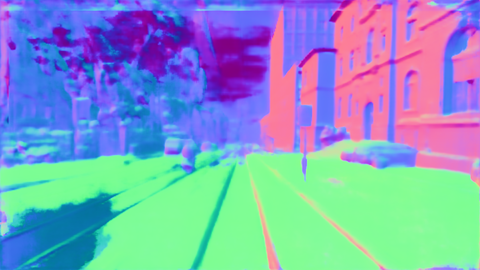}  &
\includegraphics[width=0.155\linewidth]{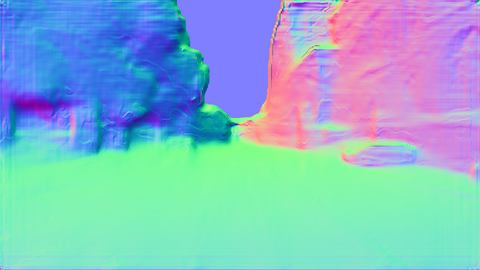}  &
\includegraphics[width=0.155\linewidth]{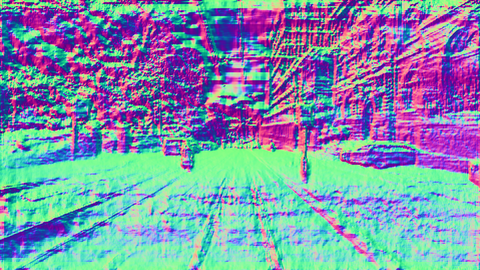}  &
\includegraphics[width=0.155\linewidth]{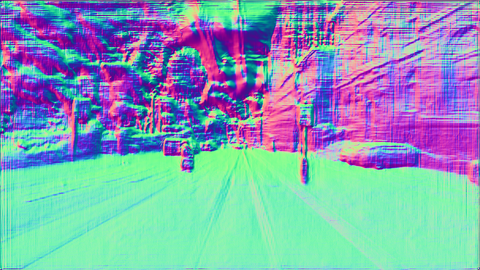}  &
\includegraphics[width=0.155\linewidth]{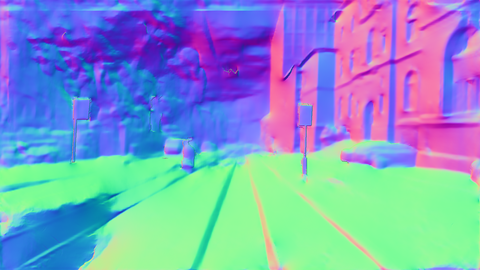}  \\
\includegraphics[width=0.155\linewidth]{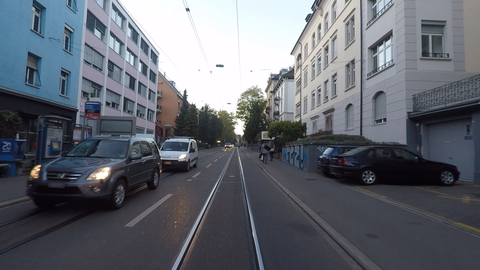}  &
\includegraphics[width=0.155\linewidth]{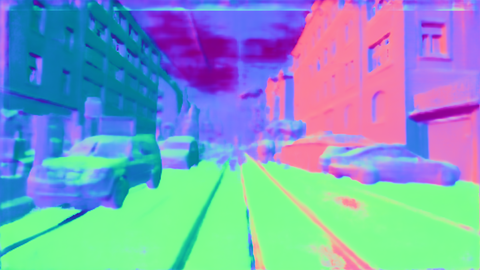}  &
\includegraphics[width=0.155\linewidth]{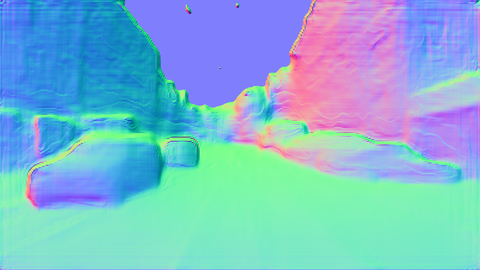}  &
\includegraphics[width=0.155\linewidth]{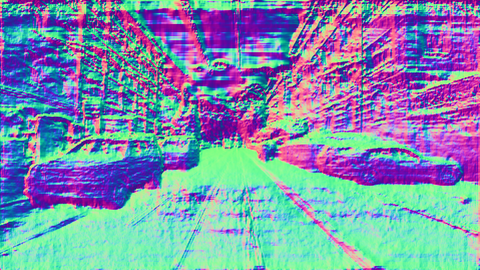}  &
\includegraphics[width=0.155\linewidth]{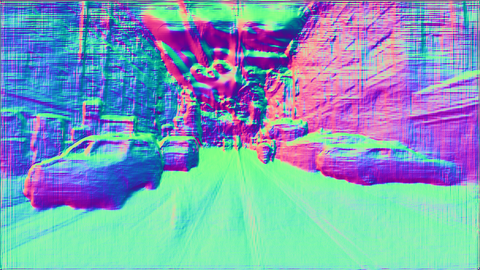}  &
\includegraphics[width=0.155\linewidth]{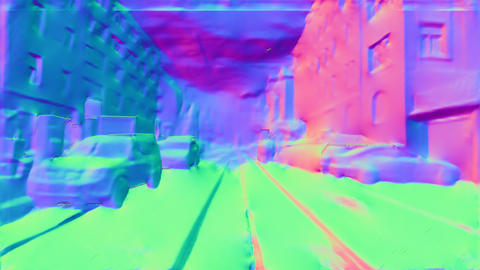}  \\
\small (a) RGB input & \small(b) iDisc~\cite{piccinelli2023idisc} & \small (c) (1) &  \small (d) (2) & \small (e) (3) & \small (f) Ours \\
\end{tabular}
    \caption{\textbf{Inferred surface normal comparison.} We compared the surface normal inferred from the reconstructed mesh with other methods (as stated in~\myfigref{fig:mesh_comp}). Our reconstructed meshes have more accurate surface normal with respect to iDisc~\cite{piccinelli2023idisc} surface normal.}
    \label{fig:normal_comparison}
\end{figure}

\begin{figure}
\centering
\begin{tabular}{@{}c@{\hspace{1mm}}c@{\hspace{1mm}}c@{\hspace{1mm}}c@{\hspace{1mm}}c@{}}
\includegraphics[width=0.24\linewidth]{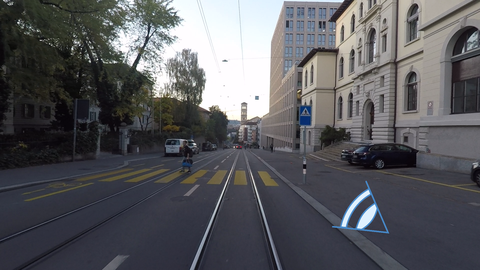}  &
\includegraphics[width=0.24\linewidth]{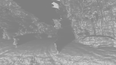}  &
\includegraphics[width=0.24\linewidth]{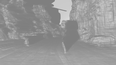}  &
\includegraphics[width=0.24\linewidth]{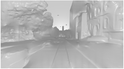}  \\
\includegraphics[width=0.24\linewidth]{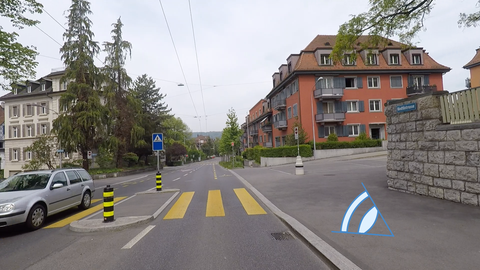}  &
\includegraphics[width=0.24\linewidth]{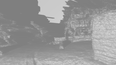}  &
\includegraphics[width=0.24\linewidth]{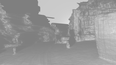}  &
\includegraphics[width=0.24\linewidth]{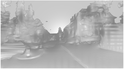}  \\
\small (a) RGB input & \small (b) (2) &  \small (c) (3) & \small (d) Ours \\
\end{tabular}
    \caption{\textbf{Mesh comparison from different angles and background mesh.} We compared the mesh reconstruction result with mesh reconstructed by other methods (, as stated in~\myfigref{fig:mesh_comp}.) with a $3$ meters zoom in and a $45$ degrees rotation clockwise as well as background mesh generated with our pipeline. The result shows that our method can efficiently break the unexpected connection between foreground and background objects and make the background mesh watertight.}
    \label{fig:mesh_angle}
\end{figure}

\begin{figure}
\label{fig:loss_curve}
\centering
\begin{tabular}{@{}c@{\hspace{1mm}}c@{\hspace{1mm}}c@{}}
\includegraphics[width=0.49\linewidth]{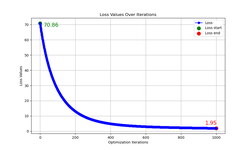}  &
\includegraphics[width=0.49\linewidth]{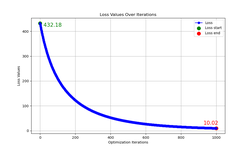}  \\
\small (a) Loss curve of upper image & \small(b) Loss curve of bottom image  \\
\end{tabular}
\vspace{-2mm}
    \caption{\textbf{Optimization loss curve.} We present the optimization loss curve of two examples shown in~\myfigref{fig:mesh_comp}.}
\end{figure}

In order to evaluate the qualitative result of our reconstructed mesh, we compare our mesh reconstruction result with results from the following settings: \textbf{(1)} mesh constructed by the Worldsheet~\cite{hu2021worldsheet} using MiDaS v2.1~\cite{Ranftl2021} as the external depth backbone based on input RGB. \textbf{(2)} mesh constructed by the Worldsheet using iDisc~\cite{piccinelli2023idisc} as the external depth backbone. \textbf{(3)} mesh reconstructed by method presented in the SIMBAR~\cite{zhang2022simbar} using Dense Prediction Transformer (DPT) monodepth models~\cite{Ranftl2022} as the external depth backbone, shown as~\myfigref{fig:mesh_comparison}. Moreover, we also compare surface normals of the generated mesh in~\myfigref{fig:normal_comparison}, further showing that our method can estimate most of the geometric characteristics and reconstruct better mesh. Noted that for surface normal, we treated estimation of iDisc~\cite{piccinelli2023idisc} as the ground truth. To show the effect of the mesh post-processing kernel, we compared our reconstructed scene mesh with  \textbf{(2)} and \textbf{(3)} from different viewing angles, as shown in~\myfigref{fig:mesh_angle}. We also present the optimization loss curve in~\myfigref{fig:loss_curve}. More qualitative results on scene mesh reconstruction are presented in~\myappendixref{apd:C}.
\subsection{Generalization to Other Datasets}
\begin{figure}[H]
\centering
\begin{tabular}{@{}c@{\hspace{1mm}}c@{\hspace{1mm}}c@{\hspace{1mm}}c@{\hspace{1mm}}c@{}}
\includegraphics[width=0.24\linewidth]{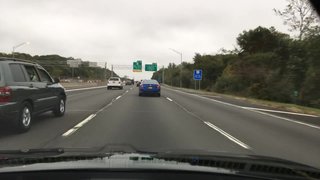}  &
\includegraphics[width=0.24\linewidth]{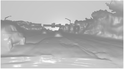}  &
\includegraphics[width=0.24\linewidth]{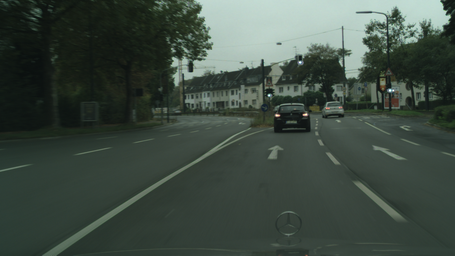}  &
\includegraphics[width=0.24\linewidth]{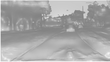}  \\
\includegraphics[width=0.24\linewidth]{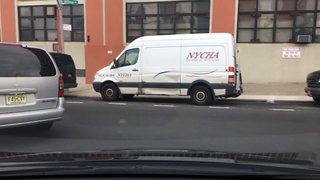}  &
\includegraphics[width=0.24\linewidth]{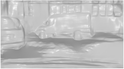}  &
\includegraphics[width=0.24\linewidth]{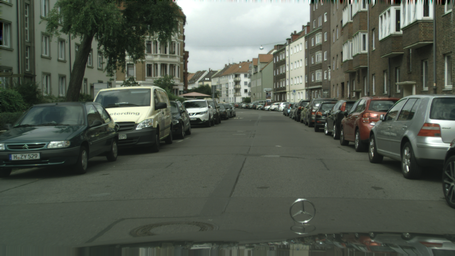}  &
\includegraphics[width=0.24\linewidth]{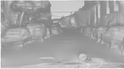}  \\
\includegraphics[width=0.24\linewidth]{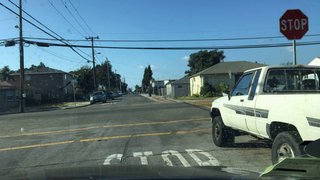}  &
\includegraphics[width=0.24\linewidth]{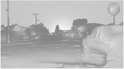}  &
\includegraphics[width=0.24\linewidth]{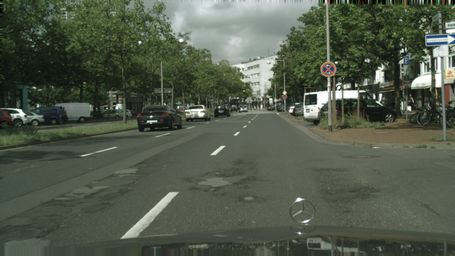}  &
\includegraphics[width=0.24\linewidth]{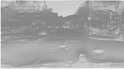}  \\
\includegraphics[width=0.24\linewidth]{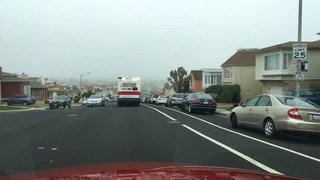}  &
\includegraphics[width=0.24\linewidth]{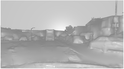}  &
\includegraphics[width=0.24\linewidth]{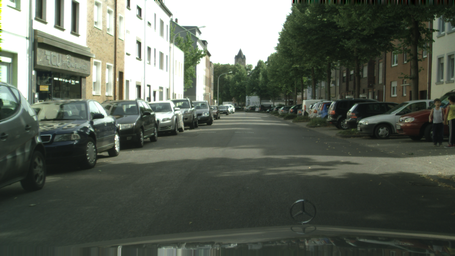}  &
\includegraphics[width=0.24\linewidth]{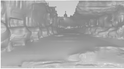}  \\
\small (a) & \small (b) &  \small (c)  & \small (d) \\
\end{tabular}
\vspace{-2mm}
\caption{\textbf{Mesh reconstructed from different datasets.} We showcase the mesh reconstruction based on two different datasets: BDD100K dataset~\cite{bdd100k} (column (a) and (b)) and Cityscapes dataset~\cite{Cordts2016Cityscapes} (column (c) and (d)). The above results demonstrate the capability of our pipeline to generate reasonable scene mesh when applied to other datasets. However, we notice that roads close to the camera are not correctly reconstructed, this is a result of inaccurate surface normal prediction caused by the ego vehicle.}
\vspace{-4mm}
\label{fig:mesh_generalize}
\end{figure}

Our pipeline can be generalized to a wide range of datasets, including the Cityscapes dataset~\cite{Cordts2016Cityscapes}, the BDD100K dataset~\cite{bdd100k} and the Dark Zurich dataset~\cite{SDV19} daytime split. To demonstrate this on the Geometric Mesh Reconstruction component, we replace the input RGB $\mathbf{I_d}$ and semantic annotation $\mathbf{S}$ with samples from the new dataset.~\myfigref{fig:mesh_generalize} illustrates the qualitative result of the scene mesh reconstructed from different datasets. We observe that our pipeline exhibits excellent generalization across various datasets.
\section{Testing Plans}
\subsection{Datasets and Metrics}
\vspace{-2mm}
\mypara{Datasets} For the experiment of the entire pipeline, we will use the ACDC dataset~\cite{SDV21ACDC} described in~\mysecref{sec:dataset}. In particular, we will apply our pipeline to reference images in which semantic annotations are available and conduct evaluation on the nighttime split testing set. The evaluation of the ACDC dataset is done via an online server. 

\mypara{Metrics} The goal of our experiment is to show that our synthetic nighttime images can serve as training data and improve the performance of current segmentation detection methods. To evaluate the retrained model, we will use the standard semantic segmentation evaluation method mean Intersection of Union (mIoU) and the Uncertainty-Aware semantic segmentation Intersection of Union (AUIoU) introduced by~\cite{SDV21ACDC}. The main distinguishing feature of those methods is the incorporation of image regions that possess indiscernible semantic content, referred as "invalid regions" during the process of annotation and evaluation.

\subsection{Experiments}
Our experiments aim to demonstrate the superiority of our generated nighttime images over the original images and other synthetic nighttime images for the task of semantic segmentation. To achieve this, we will conduct a comprehensive evaluation by training various state-of-the-art methods and architectures using different versions of nighttime images and then evaluate them using the ACDC dataset nighttime testing set. In particular, for the training data, we plan to use the original daytime reference images as the baseline, comparing it with dimmed daytime images, CycleGAN~\cite{CycleGAN2017} transferred nighttime images, our nighttime images, as well as the real nighttime images in the ACDC dataset~\cite{SDV21ACDC}. To ensure robustness and accuracy, we will train the model with various methods including DeepLabV3+~\cite{deeplabv3plus2018}, SegFormer~\cite{xie2021segformer}, Mask2Former~\cite{cheng2021mask2former}, and HRNet~\cite{SunXLW19}, as well as architectures including ResNet~\cite{He2015DeepRL}, Swin Transformer~\cite{liu2021Swin} and Vision Transformer~\cite{dosoViTskiy2020}. Moreover, for each combination listed above, we also plan to train them with different sizes of training datasets with nighttime images generated using our pipeline, showing that our pipeline is able to generate multiple nighttime images from one single daytime image by activating different light sources combinations described in~\mysecref{sec:light_active}. We will report the mIoU and AUIoU of all methods on the nighttime split of the ACDC dataset, as well as the IoU on each semantic class. Additionally, we will also present the qualitative comparison of selected semantic segmentation methods, trained using different training datasets listed above, on the ACDC dataset nighttime split. By presenting these qualitative results, we aim to better understand how different datasets impact the segmentation results and pinpoint the strengths and weaknesses of nighttime images generated using our pipeline.

%% file: text/5-conclusion.tex
%

\chapter{Conclusion}


In this thesis, we have presented a physics-based pipeline NPSim that performs day-to-night transformation with two components: Geometric Mesh Reconstruction and Realistic Nighttime Scene Relighting. This work stands apart from all prior works as it is the first to accomplish the task of relighting outdoor scenes from day to night using only a single image. The distinctiveness of our method lies in its explicit estimation of the scene's geometry and materials, and then integrate the estimated materials into the geometry alongside the light sources. The innovative aspect is the consideration of light sources that remain inactive during the daytime but become active at night. By incorporating these elements into the relighting process, this work has the potential to achieve a remarkable advancement in generating realistic night scenes from daytime photographs, setting it apart from all previous research in this field. 

Our mesh reconstruction component reconstructs better scene mesh by preserving geometric information such as depth and surface normal. It also gets rid of the potential unrealistic reflection by removing spurious faces between foreground and background objects. Moreover, the proposed photo-realistic nighttime simulation is a general approach that can be applied to any daytime driving dataset for day-to-night simulation without the need for real nighttime data. This alleviated the need to annotate large sets of real nighttime images and made a significant contribution to constituting a bottleneck for nighttime semantic scene understanding.

Meanwhile, we are also aware of several limitations of our method. Firstly, the generation of inactive light source masks is not automated and requires some human input. This problem could be alleviated by training a neural network using existing annotations to detect inactive light sources from daytime images. Secondly, our mesh reconstruction depends heavily on the estimated geometric information, errors on estimated depth and surface normal will make the reconstructed mesh inaccurate. We noticed that our current depth prediction model is not capable of correctly predicting the surface normal of some parts of the road, especially roads that are far away or under shadow. This is probably caused by domain shift between Diode dataset~\cite{diode_dataset}, NYUv2 dataset~\cite{Silberman:ECCV12} and the ACDC dataset~\cite{SDV21ACDC}. We plan to fix it by trying out different models or retraining the current model on a dataset that is closer to the ACDC dataset. Lastly, our relighting component did not consider motion blur caused by long exposure time at night, this may cause our generated nighttime images to be inconsistent with real nighttime images and lower the performance of trained models. Our future works will focus on the simulation of motion blur for moving objects and to narrow the gap between the simulated and real nighttime images.

%% file: appendix/appendix-A.tex
%
\vspace{-20mm}
\chapter{Light Source Annotation Examples}
\label{apd:A}
\vspace{-11mm}

\begin{figure}
\centering
\begin{tabular}{@{}c@{\hspace{1mm}}c@{\hspace{1mm}}c@{}}
\includegraphics[width=0.325\linewidth]{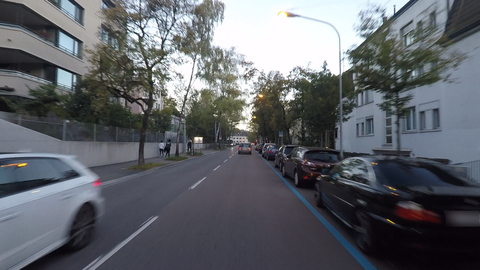} &
\includegraphics[width=0.325\linewidth]{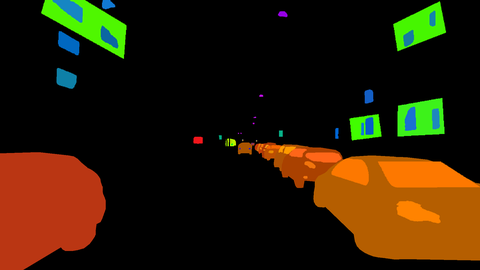}  &
\includegraphics[width=0.325\linewidth]{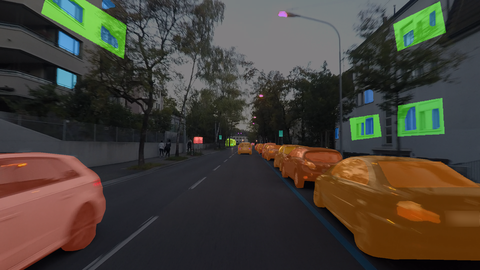}\\
\includegraphics[width=0.325\linewidth]{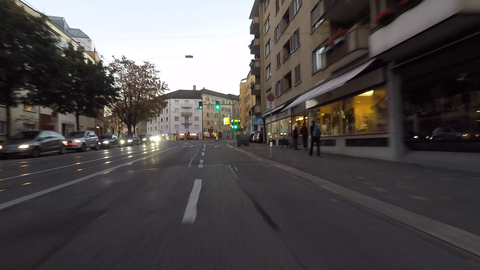} &
\includegraphics[width=0.325\linewidth]{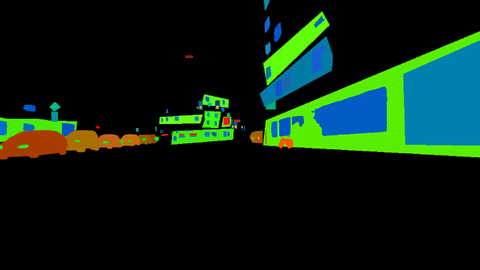}  &
\includegraphics[width=0.325\linewidth]{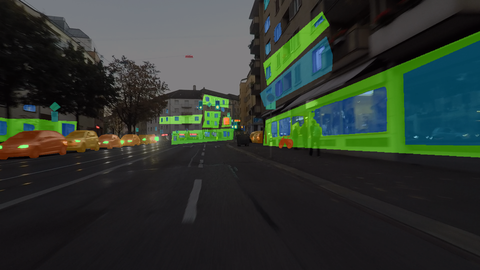}\\
\includegraphics[width=0.325\linewidth]{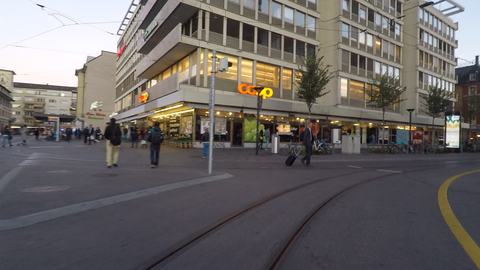} &
\includegraphics[width=0.325\linewidth]{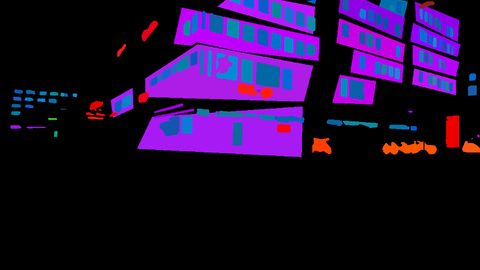}  &
\includegraphics[width=0.325\linewidth]{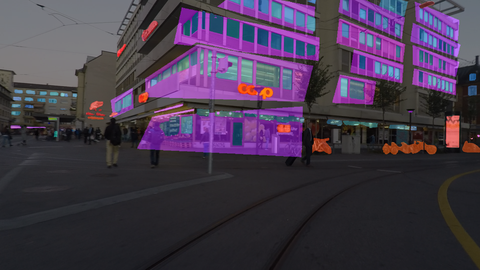}\\
\includegraphics[width=0.325\linewidth]{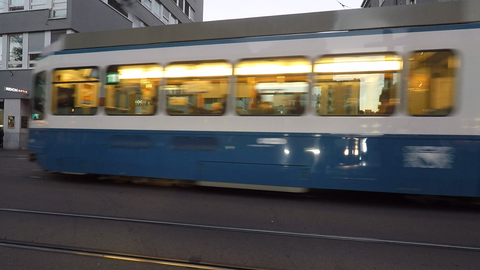} &
\includegraphics[width=0.325\linewidth]{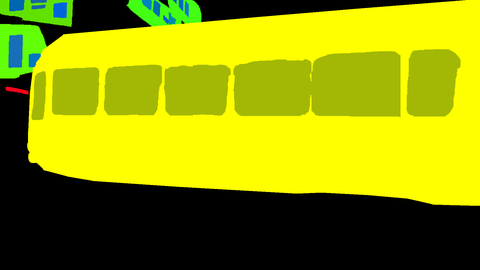}  &
\includegraphics[width=0.325\linewidth]{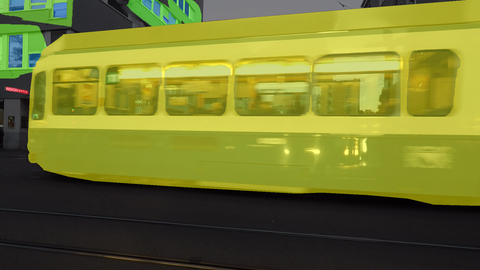}\\
\includegraphics[width=0.325\linewidth]{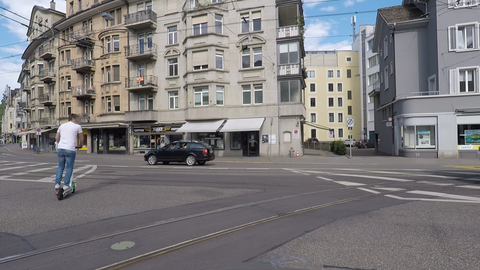} &
\includegraphics[width=0.325\linewidth]{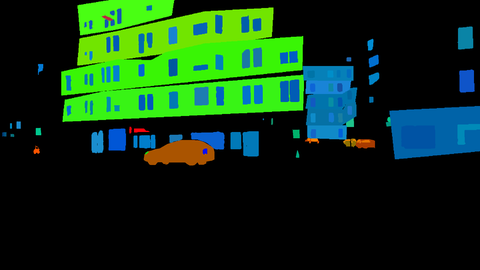}  &
\includegraphics[width=0.325\linewidth]{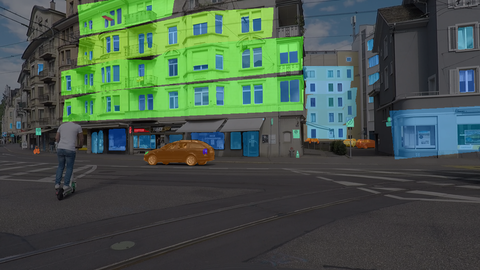}\\
\includegraphics[width=0.325\linewidth]{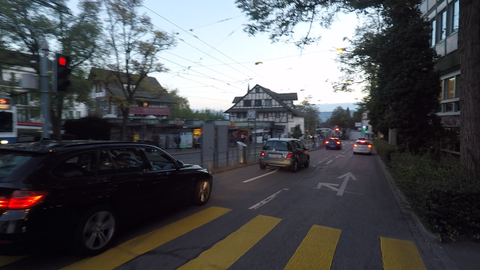} &
\includegraphics[width=0.325\linewidth]{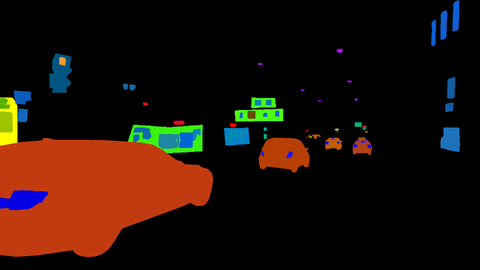}  &
\includegraphics[width=0.325\linewidth]{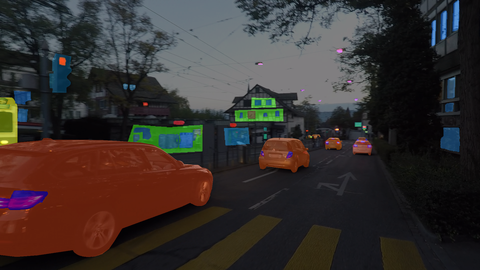}\\
\small (a) Original RGB image & \small (b) Inactive light source mask & 
\small (c) Superposition of (a) and (b)\\
\end{tabular}
\vspace{2mm}
\caption{\textbf{More Annotation Examples.} We present more examples of our inactive light source annotation. The left column shows the original RGB image, the middle column shows the annotated inactive light source mask, where each instance has its own identity and bounding box, and the right column superimposes the RGB image and the light source mask.}
\label{fig:more_annot_example}
\end{figure}

In this section, we present the appearance of each inactive light source class in both daytime and nighttime, shown in Table~\ref{tab:light_source_appearance}. Some common examples are omitted. We also provide more visual examples of our annotation, shown in~\myfigref{fig:more_annot_example}.
\nopagebreak
\begin{table}[h!]
  \centering
  \begin{tabular}{ | m{3cm} | m{6cm} | m{6cm} | }
    \hline
    Light source class & Daytime appearance & Nighttime appearance \\ \hline
    window\_building
    &
    \begin{minipage}{.3\textwidth}
    \begin{tabular}{@{}c@{\hspace{1mm}}c@{}}
      \includegraphics[height=12mm]{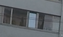}
      \includegraphics[height=12mm]{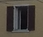}
    \end{tabular}
    \end{minipage}
    & 
    \begin{minipage}{.3\textwidth}
      \begin{tabular}{@{}c@{\hspace{1mm}}c@{}}
      \includegraphics[height=12mm]{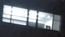}
      \includegraphics[height=12mm]{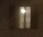}
    \end{tabular}
    \end{minipage}
    \\ \hline
    window\_parked
    &
    \begin{minipage}{.3\textwidth}
    \begin{tabular}{@{}c@{\hspace{1mm}}c@{}}
      \includegraphics[height=12mm]{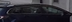}
    \end{tabular}
    \end{minipage}
    & 
    \begin{minipage}{.3\textwidth}
      \begin{tabular}{@{}c@{\hspace{1mm}}c@{}}
      \includegraphics[height=12mm]{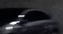}
    \end{tabular}
    \end{minipage}
    \\ \hline
    parked\_front
    &
    \begin{tabular}{@{}c@{\hspace{1mm}}c@{}}
      \includegraphics[height=12mm]{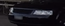}
    \end{tabular}
    & 
      \begin{tabular}{@{}c@{\hspace{1mm}}c@{}}
      \includegraphics[height=12mm]{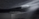}
      \includegraphics[height=12mm]{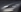}
    \end{tabular}
    \\ \hline
    parked\_rear
    &
    \begin{tabular}{@{}c@{\hspace{1mm}}c@{}}
      \includegraphics[height=12mm]{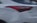}
      \includegraphics[height=12mm]{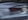}
    \end{tabular}
    & 
      \begin{tabular}{@{}c@{\hspace{1mm}}c@{}}
      \includegraphics[height=12mm]{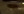}
      \includegraphics[height=12mm]{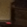}
    \end{tabular}
    \\ \hline
    window\_transport
    &
    \begin{tabular}{@{}c@{\hspace{1mm}}c@{}}
      \includegraphics[height=12mm]{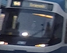}
      \includegraphics[height=12mm]{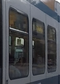}
    \end{tabular}
    & 
      \begin{tabular}{@{}c@{\hspace{1mm}}c@{}}
      \includegraphics[height=12mm]{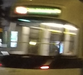}
      \includegraphics[height=12mm]{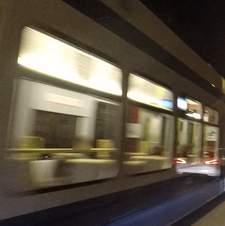}
    \end{tabular}
    \\ \hline
    street\_light\_HT
    &
    \begin{tabular}{@{}c@{\hspace{1mm}}c@{}}
      \includegraphics[height=12mm]{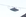}
      \includegraphics[height=12mm]{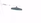}
    \end{tabular}
    & 
      \begin{tabular}{@{}c@{\hspace{1mm}}c@{}}
      \includegraphics[height=12mm]{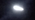}
      \includegraphics[height=12mm]{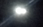}
    \end{tabular}
    \\ \hline
    street\_light\_LT
    &
    \begin{tabular}{@{}c@{\hspace{1mm}}c@{}}
      \includegraphics[height=12mm]{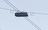}
      \includegraphics[height=12mm]{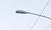}
    \end{tabular}
    & 
      \begin{tabular}{@{}c@{\hspace{1mm}}c@{}}
      \includegraphics[height=12mm]{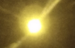}
      \includegraphics[height=12mm]{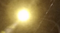}
    \end{tabular}
    \\ \hline
    advertisement
    &
    \begin{tabular}{@{}c@{\hspace{1mm}}c@{}}
      \includegraphics[height=12mm]{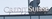}
      \includegraphics[height=12mm]{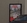}
    \end{tabular}
    & 
      \begin{tabular}{@{}c@{\hspace{1mm}}c@{}}
      \includegraphics[height=12mm]{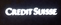}
      \includegraphics[height=12mm]{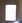}
    \end{tabular}
    \\ \hline
    clock
    &
    \begin{tabular}{@{}c@{\hspace{1mm}}c@{}}
      \includegraphics[height=12mm]{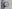}
    \end{tabular}
    & 
      \begin{tabular}{@{}c@{\hspace{1mm}}c@{}}
      \includegraphics[height=12mm]{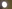}
    \end{tabular}
    \\ \hline
    inferred
    &
    \begin{tabular}{@{}c@{\hspace{1mm}}c@{}}
      \includegraphics[height=12mm]{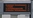}
    \end{tabular}
    & 
      \begin{tabular}{@{}c@{\hspace{1mm}}c@{}}
      \includegraphics[height=12mm]{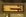}
    \end{tabular}
    \\ \hline
  \end{tabular}
  \vspace{-1mm}
  \caption{\textbf{Inactive light sources examples.} We present examples of different types of inactive light sources. The second and third columns show the daytime appearance and nighttime appearance of each example, respectively, arranged in the same order.}\label{tbl:myLboro}
  \label{tab:light_source_appearance}
\end{table}

%% file: appendix/appendix-B.tex
\chapter{Hyper-parameter Tuning}
\label{apd:B}
To find the optimal choice of hyper-parameters used in the depth refinement kernel, we apply grid search to decide $\lambda_1 \sim \lambda_3$ in~\myeqref{equ:final_loss}. In our experiment, we set $\lambda_1$ to $1$ and then changing $\lambda_2$ and $\lambda_3$ between $10^{-3}$ and $10^3$.In particular, we select $7$ values for each parameter: $10^{-3}$, $10^{-2}$, $10^{-1}$, $10^0$, $10^1$, $10^2$, $10^3$. We tried out 49 different combinations in total. Next, we will present the effect of each loss term on mesh reconstruction by showing qualitative results of reconstructed mesh and optimized surface normal.

\begin{figure}[h!]
\centering
\begin{tabular}{@{}c@{\hspace{1mm}}c@{\hspace{1mm}}c@{}}
\includegraphics[width=0.325\linewidth]{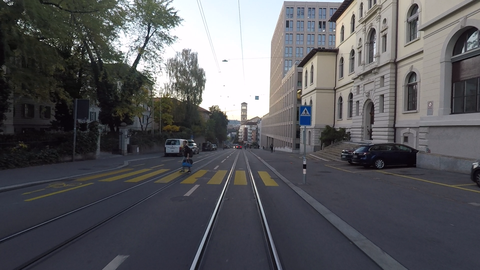} &
\includegraphics[width=0.325\linewidth]{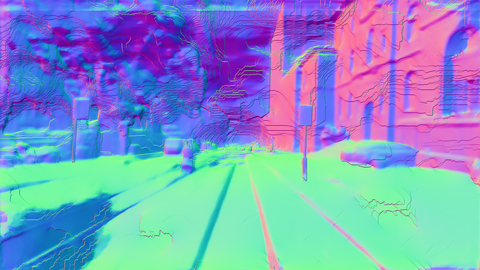}  &
\includegraphics[width=0.325\linewidth]{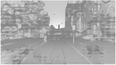}\\
\includegraphics[width=0.325\linewidth]{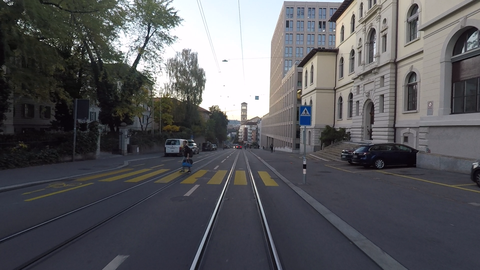} &
\includegraphics[width=0.325\linewidth]{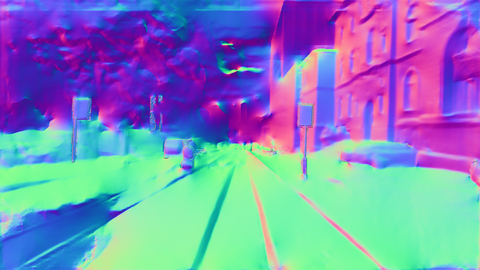}  &
\includegraphics[width=0.325\linewidth]{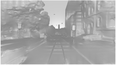}\\
\small (a) Original RGB image & \small (b) Refined surface normal & 
\small (c) Reconstructed mesh\\
\end{tabular}
\caption{\textbf{Small $\lambda_2$.} We present an example of surface normal and final mesh with different $\lambda_2$ values. The upper row shows an example of setting $\lambda_2$ too small and the lower row shows an example of setting $\lambda_2$ too large.}
\label{fig:small_con_loss}
\end{figure}

During our experiments, the coefficient of continuity loss $\lambda_2$ plays a more important role. When setting $\lambda_2$ small, gaps caused by sudden changes in depth will be created in both refined surface normal and reconstructed mesh. Those gaps will not violate other loss terms much but will have a significant negative impact on relighting. The upper row of ~\myfigref{fig:small_con_loss} shows an example of gaps on surface normals and reconstructed mesh. Similarly, having $\lambda_2$ too large will also cause some negative effects, as shown in the lower row of ~\myfigref{fig:small_con_loss}. Though it doesn't have a negative effect on the refined surface normal, the final reconstructed mesh will have some wave-like artifacts, this kind of effect is mainly caused by the dual-reference cross-bilateral filter during depth refinement.

\begin{figure}
\centering
\begin{tabular}{@{}c@{\hspace{1mm}}c@{\hspace{1mm}}c@{}}
\includegraphics[width=0.325\linewidth]{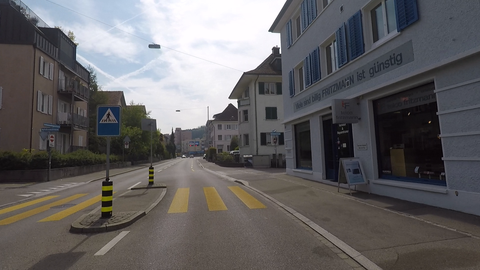} &
\includegraphics[width=0.325\linewidth]{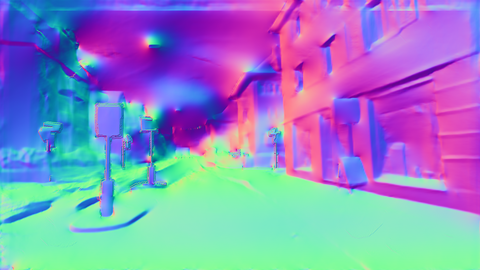}  &
\includegraphics[width=0.325\linewidth]{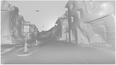}\\
\small (a) Original RGB image & \small (b) Refined surface normal & 
\small (c) Reconstructed mesh\\
\end{tabular}
\vspace{-2mm}
\caption{\textbf{Large $\lambda_3$.} We present an example of surface normal and final mesh when setting $\lambda_3$ too large.}
\label{fig:large_depth_loss}
\end{figure}

Compared to $\lambda_2$, $\lambda_3$ (the coefficient of depth loss) plays a less important role. We discover that making $\lambda_3$ too large will not affect the reconstruction result much. However, having it too small will make the refined depth deviate more from the originally predicted depth, especially for foreground objects as they are usually surrounded by uncertain regions. This will further result in wider uncertain regions and remaining unexpected faces after after mesh post-processing component. We present one visual example in the~\myfigref{fig:large_depth_loss}.

Similarly, the effect of $\lambda_1$ (the coefficient of normal loss) will largely depend on $\lambda_2$ and $\lambda_3$ together. When setting both $\lambda_2$ and $\lambda_3$ large, the result will have a combined effect of large $\lambda_2$ and large $\lambda_3$, with wider uncertain region and wave-like visual artifacts. When setting both $\lambda_2$ and $\lambda_3$ smaller, we observed similar effects as shown in the upper row of~\myfigref{fig:small_con_loss}. In the end, we set $\lambda_1 = 1$, $\lambda_2 = 5$ and $\lambda_3 = 1$

%% file: appendix/appendix-C.tex
\chapter{Mesh Reconstruction Examples}
\label{apd:C}
In this section, we present more qualitative examples of the reconstructed mesh and its corresponding surface normal, shown in the figure below. Mesh reconstructed using our method has smoother surface normal and more detailed geometry information.

\begin{figure}[h!]
\centering
\begin{tabular}{@{}c@{\hspace{1mm}}c@{\hspace{1mm}}c@{\hspace{1mm}}c@{}}
\includegraphics[width=0.325\linewidth]{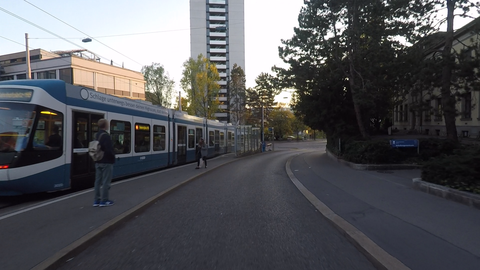}  &
\includegraphics[width=0.325\linewidth]{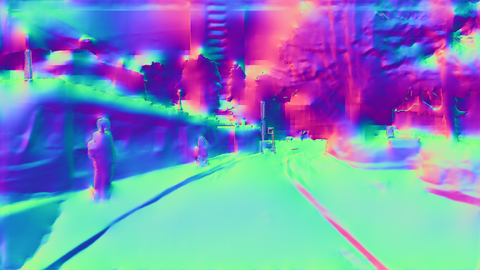}  &
\includegraphics[width=0.325\linewidth]{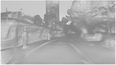} \\
\includegraphics[width=0.325\linewidth]{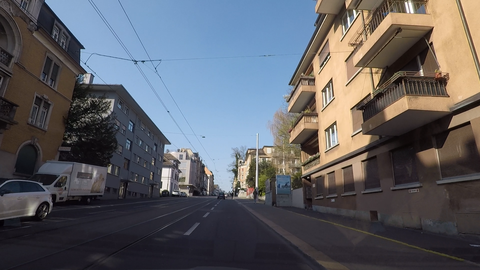}  &
\includegraphics[width=0.325\linewidth]{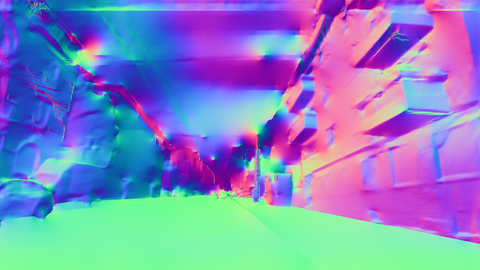}  &
\includegraphics[width=0.325\linewidth]{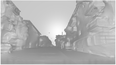} \\
\includegraphics[width=0.325\linewidth]{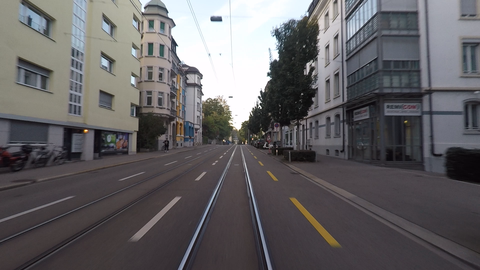}  &
\includegraphics[width=0.325\linewidth]{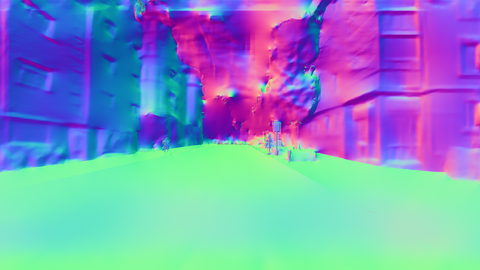}  &
\includegraphics[width=0.325\linewidth]{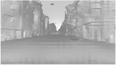} \\
\includegraphics[width=0.325\linewidth]{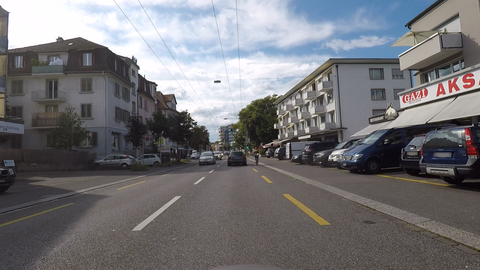}  &
\includegraphics[width=0.325\linewidth]{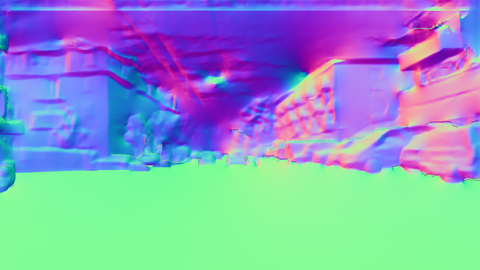}  &
\includegraphics[width=0.325\linewidth]{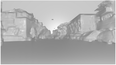} \\
\small (a) RGB input & \small(b) Corresponding surface normal & \small (c) Reconstructed mesh  \\
\end{tabular}
    \caption{\textbf{More mesh reconstruction examples.} We present more examples of mesh reconstructed using our method and their corresponding surface normal.}
    \label{fig:more_normal_comparison}
\end{figure}

\begin{figure}
\centering
\begin{tabular}{@{}c@{\hspace{1mm}}c@{\hspace{1mm}}c@{\hspace{1mm}}c@{}}
\includegraphics[width=0.325\linewidth]{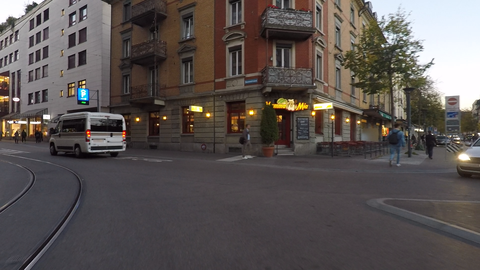}  &
\includegraphics[width=0.325\linewidth]{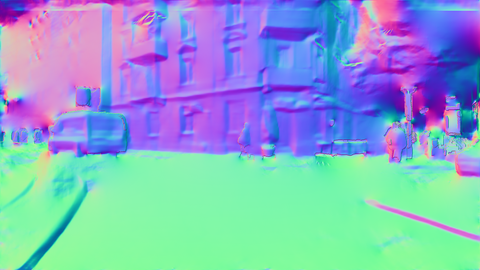}  &
\includegraphics[width=0.325\linewidth]{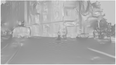} \\
\includegraphics[width=0.325\linewidth]{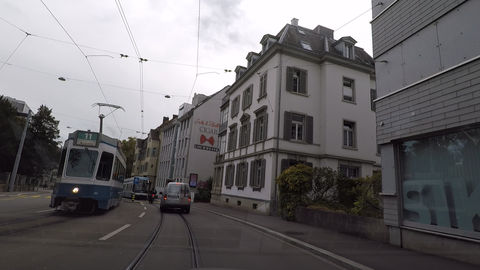}  &
\includegraphics[width=0.325\linewidth]{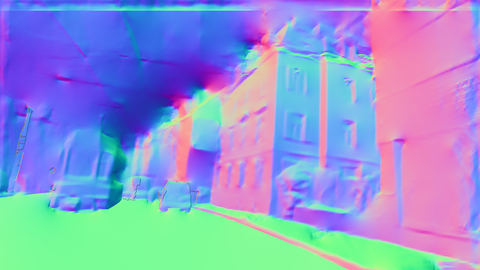}  &
\includegraphics[width=0.325\linewidth]{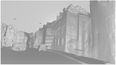} \\
\includegraphics[width=0.325\linewidth]{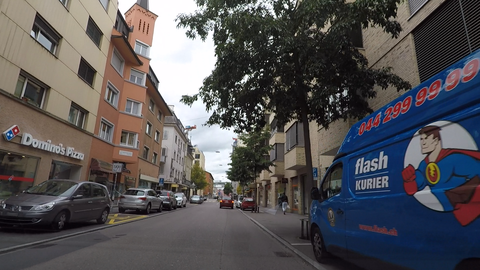}  &
\includegraphics[width=0.325\linewidth]{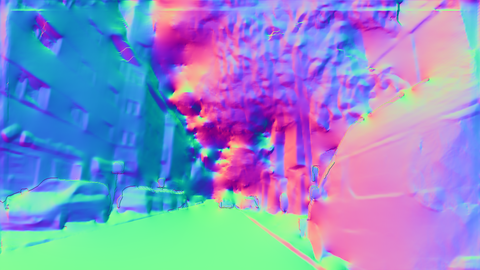}  &
\includegraphics[width=0.325\linewidth]{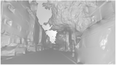} \\
\includegraphics[width=0.325\linewidth]{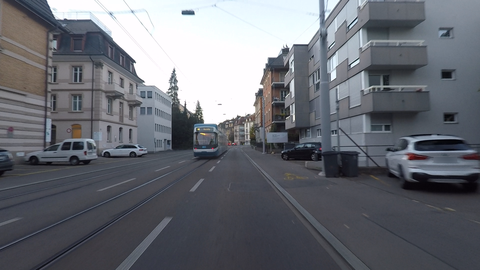}  &
\includegraphics[width=0.325\linewidth]{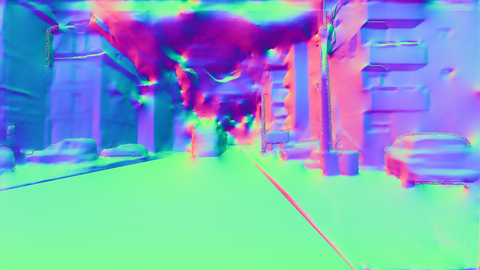}  &
\includegraphics[width=0.325\linewidth]{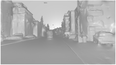} \\
\includegraphics[width=0.325\linewidth]{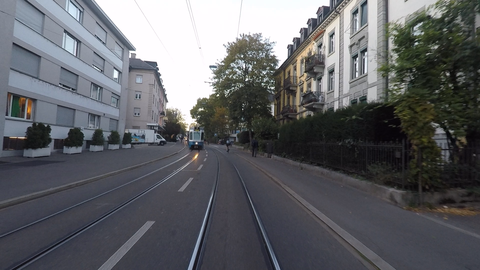}  &
\includegraphics[width=0.325\linewidth]{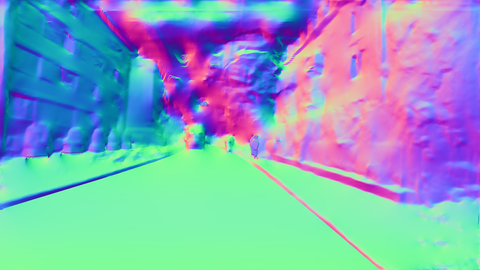}  &
\includegraphics[width=0.325\linewidth]{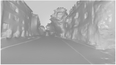} \\
\includegraphics[width=0.325\linewidth]{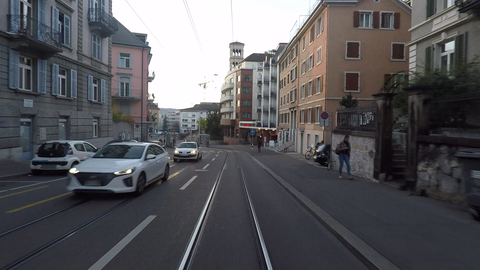}  &
\includegraphics[width=0.325\linewidth]{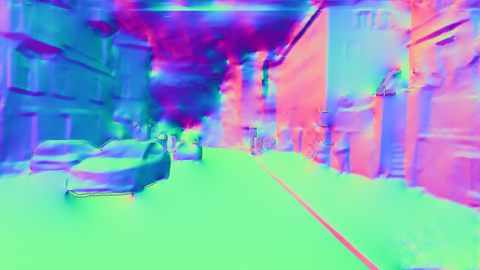}  &
\includegraphics[width=0.325\linewidth]{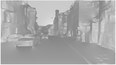} \\
\small (a) RGB input & \small(b) Corresponding surface normal & \small (c) Reconstructed mesh  \\
\end{tabular}
    \caption{\textbf{More mesh reconstruction examples.} We present more examples of mesh reconstructed using our method and their corresponding surface normal.}
    \label{fig:more_normal_comparison_more}
\end{figure}

%% file: appendix/appendix-D.tex
\chapter{Model Selection}
\label{apd:E}

In this section, we present the selection process of network $\mathbf{F_g}$ and $\mathbf{F_ir}$. For network $F_g$, we directly use the iDisc depth model pre-trained on the KITTI dataset~\cite{Geiger2012CVPR}. However, for the surface normal prediction, as for our own knowledge, all previous works have focused on indoor datasets such as the NYUv2 dataset~\cite{Silberman:ECCV12}. To overcome the domain gap between indoor and outdoor scenes, we retrained the iDisc surface normal network on the Diode dataset~\cite{diode_dataset} outdoor split. To maximize the performance, we tried several training settings, with their evaluation results shown in Table~\ref{tab:normal_result}.

\begin{table*}[h!]
\newcolumntype{Z}{S[table-format=2.3,table-auto-round]}
\centering
\vspace{-0.5em}
\setlength{\tabcolsep}{3mm}
\small
\footnotesize
\centering
\begin{tabular}{|c|c|c|c|c|c|}\hline
Setting & rmse\_angular & $a_1$($<$5deg) & $a_2$($<$11.5deg) & $a_3$($<$22.5deg) & $a_4$($<$30deg) \\\hline
iDisc NYUv2 & 60.181 & 0.037& 0.184 & 0.312 & 0.387 \\\hline
iDisc NYUv2+Diode & 77.862 & 0.044 & 0.130 & 0.212 & 0.281 \\\hline
iDisc Diode & \textbf{44.874} & \textbf{0.290} & \textbf{0.435} & \textbf{0.563} & \textbf{0.625} \\\hline
           
\end{tabular}
\caption{{\bf iDisc normal estimation retrain result}. We present three different settings of our networks and their corresponding evaluation result.}
\label{tab:normal_result}
\end{table*}